\pdfoutput=1

\documentclass[11pt]{article}

\usepackage{acl}

\usepackage{times}
\usepackage{latexsym}

\usepackage{times}
\usepackage{latexsym}
\usepackage{CJKutf8}
\usepackage{amsmath}
\usepackage{verbatim}
\usepackage{booktabs}
\usepackage{amssymb}
\usepackage{pifont}
\usepackage{stfloats}
\usepackage{multicol}
\usepackage{float}
\usepackage{graphicx}
\usepackage{subfigure}
\usepackage{multirow}
\usepackage{array}
\newcommand{\PreserveBackslash}[1]{\let\temp=\\#1\let\\=\temp}
\newcommand{\bm}[1]{\mbox{\boldmath{$#1$}}}
\newcolumntype{C}[1]{>{\PreserveBackslash\centering}p{#1}}
\newcolumntype{R}[1]{>{\PreserveBackslash\raggedleft}p{#1}}
\newcolumntype{L}[1]{>{\PreserveBackslash\raggedright}p{#1}}
\usepackage{diagbox}
\usepackage[noend]{algpseudocode}
\usepackage{algorithmicx,algorithm}
\usepackage{arydshln}
\usepackage{booktabs}
\usepackage{amsfonts}
\usepackage{bbm}
\usepackage{mathtools}

\makeatletter
\def\hlinew#1{%
  \noalign{\ifnum0=`}\fi\hrule \@height #1 \futurelet
   \reserved@a\@xhline}
\makeatother
\usepackage[T1]{fontenc}

\usepackage[utf8]{inputenc}

\usepackage{microtype}

\title{Modeling Dual Read/Write Paths for Simultaneous Machine Translation}

\author{Shaolei Zhang \textsuperscript{\rm 1,2},
    Yang Feng \textsuperscript{\rm 1,2}\thanks{ $\;\;$Corresponding author: Yang Feng. $\;\;\;\;\;\;\;\;\;\;\;\;\;\;\;\;$ Code is available at: \url{https://github.com/ictnlp/Dual-Path}} \\
        \textsuperscript{\rm 1}{Key Laboratory of Intelligent Information Processing} \\ Institute of Computing Technology, Chinese Academy of Sciences (ICT/CAS) \\
    { \textsuperscript{\rm 2} {University of Chinese Academy of Sciences, Beijing, China}} \\
     \texttt{\{zhangshaolei20z, fengyang\}@ict.ac.cn}  }

\begin{document}
\maketitle
\begin{abstract}

Simultaneous machine translation (SiMT) outputs translation while reading source sentence and hence requires a policy to decide whether to wait for the next source word (READ) or generate a target word (WRITE), the actions of which form a \emph{read/write path}. Although the read/write path is essential to SiMT performance, no direct supervision is given to the path in the existing methods. In this paper, we propose a method of dual-path SiMT which introduces duality constraints to direct the read/write path. According to duality constraints, the read/write path in source-to-target and target-to-source SiMT models can be mapped to each other. As a result, the two SiMT models can be optimized jointly by forcing their read/write paths to satisfy the mapping. Experiments on En$\leftrightarrow$Vi and De$\leftrightarrow$En tasks show that our method can outperform strong baselines under all latency.

\end{abstract}

\section{Introduction}

Simultaneous machine translation (SiMT) \cite{Cho2016,gu-etal-2017-learning,ma-etal-2019-stacl,Arivazhagan2019}, which outputs translation while reading source sentence, is important to many live scenarios, such as simultaneous interpretation, live broadcast and synchronized subtitles. Different from full-sentence machine translation which waits for the whole source sentence, SiMT has to decide whether to wait for the next source word (i.e., READ action) or translate a target word (i.e., WRITE action) to complete the translation. 

\begin{figure}[t]
\centering
\subfigure[Segment pairs between the sentence pair.]{
\includegraphics[width=3in]{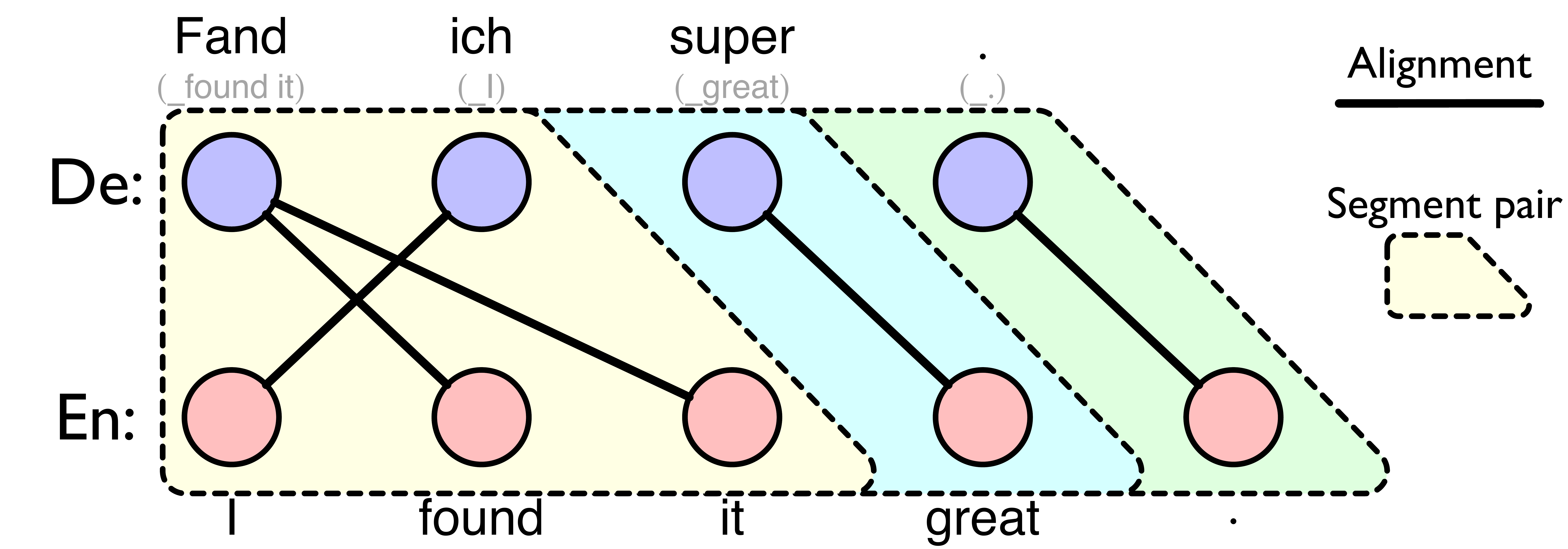}
\label{ill1}
}
\subfigure[The duality between the read/write paths in two directions.]{
\includegraphics[width=3.0in]{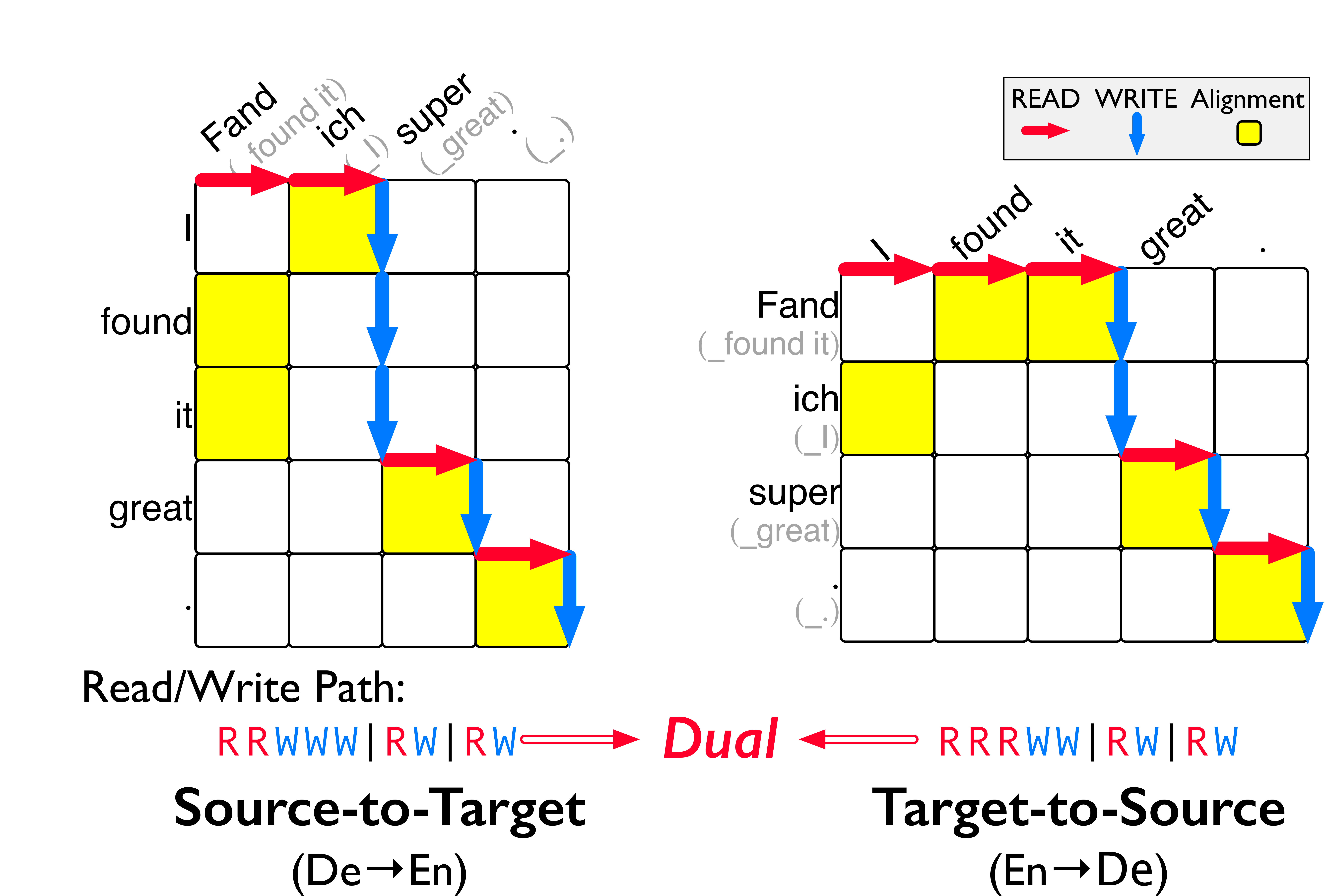}
\label{ill2}
}

\caption{An example of duality constraints. With duality constraints, the read/write paths of source-to-target and target-to-source translation should project to the same segment pairs between two languages.}
\label{ill}
\end{figure}

The sequence of READ and WRITE actions in the translation process form a \emph{read/write path}, which is key to SiMT performance. Improper read/write path will bring damage to translation performance as compared to the following WRITE actions too many but not \emph{necessary} READ actions will result in high translation latency while too few but not \emph{sufficient} READ actions will exclude indispensable source information. Therefore, an ideal read/write path is that the READ actions compared to the following WRITE actions are just \emph{sufficient} and \emph{necessary}, which means the source words covered by consecutive READ actions and the target words generated by the following consecutive WRITE actions should be semantically equivalent. 

Ensuring sufficiency and necessity between READ/WRITE actions will lead to a proper read/write path and thereby good SiMT performance. But unfortunately, the existing SiMT methods, which employ a fixed or adaptive policy, do not consider the sufficiency or necessity in their policy. The fixed policy performs SiMT based on a pre-defined read/write path \cite{dalvi-etal-2018-incremental, ma-etal-2019-stacl}, where the number of READ actions before WRITE is fixed. The adaptive policy \cite{gu-etal-2017-learning,Zheng2019a,Arivazhagan2019,Zheng2019b, Ma2019a,liu-etal-2021-cross} dynamically decides to READ or WRITE guided by translation quality and total latency, but skips the evaluation of sufficiency and necessity between READ/WRITE actions. 

Under these grounds, we aim at introducing the evaluation of sufficiency and necessity between READ/WRITE actions to direct the read/write path without involving external information. As mentioned above, in an ideal read/write path, the source segment (i.e., source words read by the consecutive READ actions) and the corresponding target segment (i.e., target words generated by the following consecutive WRITE actions) are supposed to be semantically equivalent and thus translation to each other, which constitutes a separate \emph{segment pair}. Hence, an ideal read/write path divides the whole sentence pair into a sequence of segment pairs where the source sentence and the target sentence should be translation to each other segment by segment. That means if the translation direction is reversed, an ideal read/write path for target-to-source SiMT can also be deduced from the same sequence of segment pairs. For example, according to the alignment in Figure \ref{ill1}, the ideal read/write paths should be `\textrm{RRWWW$\mid$RW$\mid$RW}' in De$\rightarrow$En SiMT and `\textrm{RRRWW$\mid$RW$\mid$RW}' in En$\rightarrow$De SiMT, as shown in Figure \ref{ill2}, both of which share the same segment pair sequence of $\left<\textit{Fand ich}, \textit{I fount it} \right>$, $\left<\textit{super}, \textit{great}\right>$ and $\left<\textit{.}, \textit{.}\right>$. Therefore, agreement on the segment pairs derived from read/write paths in source-to-target and target-to-source SiMT, called \emph{duality constraints}, can be a good choice to evaluate sufficiency and necessity between READ/WRITE actions.

Based on the above reasoning, we propose a method of \emph{Dual-Path SiMT}, which uses the SiMT model in the reverse direction to guide the SiMT model in the current direction according to duality constraints between their read/write paths. With duality constraints, the read/write paths in source-to-target and target-to-source SiMT should reach an agreement on the corresponding segment pairs. Along this line, our method maintains a source-to-target SiMT model and a target-to-source SiMT model concurrently, which respectively generate their own read/write path using monotonic multi-head attention \cite{Ma2019a}. By minimizing the difference between the segment pairs derived from the two read/write paths, the two SiMT models successfully converge on the segment pairs and provide supervision to each other. Experiments on IWSLT15 En$\leftrightarrow$Vi and WMT15 De$\leftrightarrow$En SiMT tasks show that our method outperforms strong baselines under all latency, including the state-of-the-art adaptive policy. 

\section{Background}
\label{sec:Monotonic attention}
We first briefly introduce SiMT with a focus on monotonic multi-head attention \cite{Ma2019a}.

For a SiMT task, we denote the source sentence as $\mathbf{x}\!=\!\left \{ x_{1},\cdots ,x_{J} \right \}$ and the corresponding source hidden states as $\mathbf{m}\!=\!\left \{ m_{1},\cdots ,m_{J} \right \}$, where $J$ is the source length. The model generates target sentence $\mathbf{y}\!=\!\left \{ y_{1},\cdots ,y_{I} \right \}$ with target hidden states $\mathbf{s}\!=\!\left \{ s_{1},\cdots ,s_{I} \right \}$, where $I$ is the target length. During translating, SiMT model decides to read a source word (READ) or write a target word (WRITE) at each step, forming a read/write path. 

\textbf{Read/write path} can be represented in multiple forms, such as an action sequence of READ and WRITE (e.g., RRWWWRW$\cdots$), or a path from $(0,0)$ to $(I,J)$ in the attention matrix from the target to source, as shown in Figure \ref{ill2}. 

Mathematically, a read/write path can be represented by a monotonic non-decreasing sequence $\left \{ g_{i} \right \}_{i=1}^{I}$ of step $i$, where the $g_{i}$ represents the number of source words read in when writing the $i^{th}$ target word $y_{i}$. The value of $\left \{ g_{i} \right \}_{i=1}^{I}$ depends on the specific SiMT policy, where monotonic multi-head attention (MMA) \cite{Ma2019a} is the current state-of-the-art SiMT performance via modeling READ/WRITE action as a Bernoulli variable.

\textbf{Monotonic multi-head attention} MMA processes the source words one by one, and concurrently predicts a selection probability $p_{ij}$ to indicates the probability of writing $y_{i}$ when reading $x_{j}$, and accordingly a Bernoulli random variable $z_{ij}$ is calculated to determine READ or WRITE action: 
\begin{align}
    p_{ij}=&\mathrm{Sigmoid}\left (\frac{m_{j}V^{K}(s_{i-1}V^{Q})^{\top }}{\sqrt{d_{k}}} \right), \label{eq2} \\
    z_{ij}\sim&\; \mathrm{Bernoulli}\left(p_{ij} \right), \label{eq3}
\end{align}
where $V^{K}$ and $V^{Q}$ are learnable parameters, $d_{k}$ is dimension of head. If $z_{ij}=0$, MMA performs READ action to wait for next source word. If $z_{ij}=1$, MMA sets $g_{i}=j$ and performs WRITE action to generate $y_{i}$ based on $x_{\leq g_{i}}$. Therefore, the decoding probability of $\mathbf{y}$ with parameters $\bm{\theta }$ is
\begin{gather}
    p(\mathbf{y}\mid \mathbf{x}; \bm{\theta })=\prod_{i=1}^{I}p\left ( y_{i}\mid \mathbf{x}_{\leq g_{i}},\mathbf{y}_{< i} ; \bm{\theta }\right ),
\end{gather}
where $\mathbf{x}_{\leq g_{i}} $ are first $g_{i}$ source tokens, and $\mathbf{y}_{< i}$ are previous target tokens. 

Note that when integrated into multi-head attention, all attention heads in decoder layers independently determine the READ/WRITE action. If and only when all heads decide to perform WRITE action, the model starts translating, otherwise the model waits for the next source word.

\textbf{Expectation training} Since sampling a discrete random variable $z_{ij}$ precludes back-propagation, MMA applies expectation training \citet{LinearTime} to replace $z_{ij}$ with a \emph{expected writing probability}, denoted as
\begin{gather}
\bm{\alpha}=(\alpha_{ij})_{I\times J}, 
\label{eq-alpha}
\end{gather}
where $\alpha_{ij}$ calculates the expectation probability of writing $y_{i}$ when reading $x_{j}$. Then, the attention distribution and context vectors are accordingly calculated in the expected form.

To trade-off between translation quality and latency, MMA introduces a latency loss $\mathcal{L}_{g}$ to the total loss function:
\begin{equation}
    \mathcal{L}\left ( \bm{\theta } \right )=-\sum_{(\mathbf{x},\mathbf{y})}\mathrm{log}\; p\left (\mathbf{y}\mid \mathbf{x}; \bm{\theta } \right )+\lambda \mathcal{L}_{g}, \label{eq15}
\end{equation}
where $\mathcal{L}_{g}$ measures the total latency, and $\lambda$ is the weight of latency loss. Please refer to \citet{Arivazhagan2019} and \citet{Ma2019a} for more detailed derivation and implementation.

\section{The Proposed Method}
\label{sec:method}

\begin{figure}[t]
\centering
\includegraphics[width=3.05in]{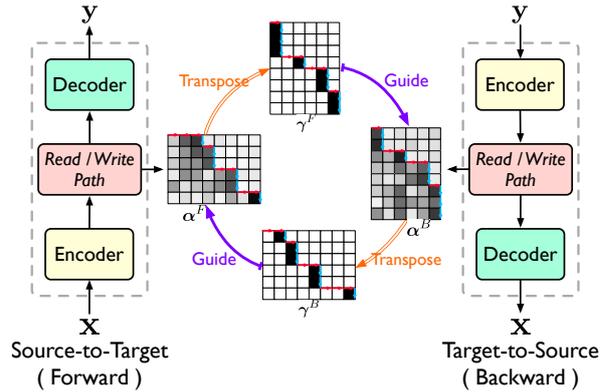}
\caption{The architecture of dual-path SiMT, consisting of the forward and backward single-path SiMT models. To accomplish the duality constraints, we generate the transposed path of the forward (or backward) read/write path, and use this transposed path to direct the read/write path in another direction; vice versa.}
\label{model}
\end{figure}

Our dual-path SiMT model employs a source-to-target (\emph{forward}) model and a target-to-source (\emph{backward}) model, called single-path SiMT, which generate their own read/write path based on MMA. According to duality constraints that the read/write paths of the two single-path SiMT models should share the same segment pair sequence, the two read/write paths can be transposed to each other in principle as shown in Figure \ref{ill}. But in practice, there is a gap between the  transposed read/write path and the original one in the reverse translation direction. By closing the gap between the aforementioned transposed and original read/write paths, as shown in Figure \ref{model}, duality constraints are introduced into the dual-path SiMT model and thereby the two single-path SiMT models can provide guidance to each other. In what follows, we will introduce how to get the transposed read/write path (Sec.\ref{sec:tp}) and how to reduce the gap (Sec.\ref{sec:dualpath}).

\subsection{Transposing the Read/Write Path}
\label{sec:tp}

\begin{figure*}[t]
\centering
\includegraphics[width=6.23in]{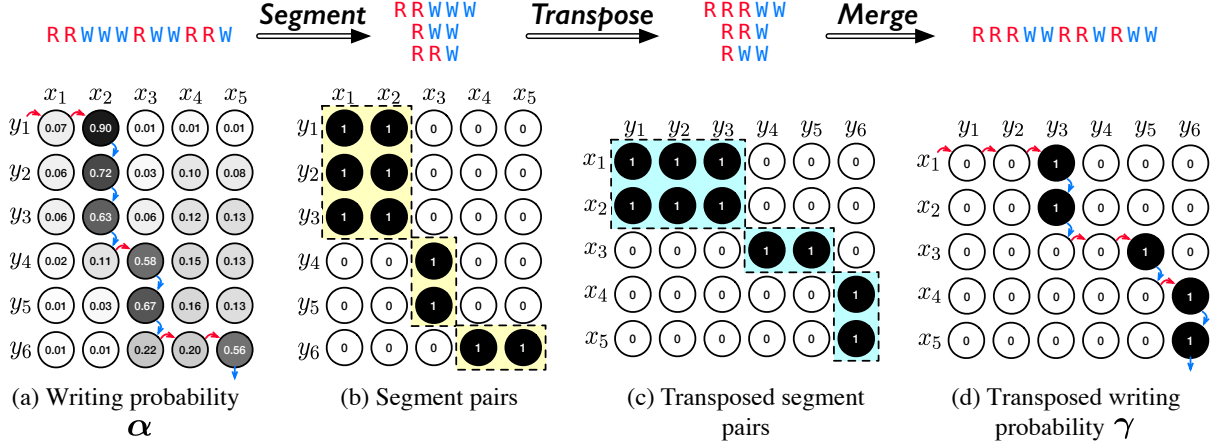}
\caption{Simplified diagrams of generating transposed writing probability $\bm{\gamma}$ from the writing probability $\bm{\alpha}$. (a$\rightarrow$b) Segment the sentence pair into a sequence of segment pairs. (b$\rightarrow$c) Transpose the segment pairs to fit the backward SiMT. (c$\rightarrow$d) Merge the transposed segment pairs to get transposed writing probability with the transposed path.}
\label{attn}
\end{figure*}

The purpose of transposing a read/write path is to get a new read/write path in the reverse direction based on the same segment pairs as the original path. As the transposing process works in the same way for the two directions, we just introduce the process for the forward single-path SiMT. Since there is no explicit read/write path in the training of single-path SiMT model, the transposing process can only use the expected writing probability matrix $\bm{\alpha}$ as the input, shown in Eq.(\ref{eq-alpha}). Similarly, the output of the transposing process is the transposed writing probability matrix $\bm{\gamma}$ calculated from the transposed read/write path, which will be used to guide the backward single-path SiMT.

The transposing process consists of three steps. First, derive the read/write path from the expected writing probability matrix $\bm{\alpha}$ and \emph{segment} the sentence pair into a sequence of segment pairs. Second, \emph{transpose} the sequence of segment pairs into the corresponding one for the backward SiMT. Last, \emph{merge} the transposed segment pairs to get the transposed path and then project it to $\bm{\gamma}$. In the following, we will introduce the steps of segment, transpose and merge in details.

\textbf{Segment} Given the expected writing probability matrix $\bm{\alpha}$, to get the read/write path, we first find out the source position $d_i$ that the WRITE action for each target position $i$ corresponds to, which is 
\begin{gather}
    d_{i}=\underset{j}{\mathrm{argmax}}\;\alpha_{ij}. \label{eq16}
\end{gather}
According to the property of monotonic attention, there are some consecutive WRITE actions that corresponds to the same source position, so the target words generated by the consecutive WRITE actions form a target segment. Formally, assuming there are $K$ target segments in total, denoted as $\mathbf{y}\!=\!\left \{ \bar{\mathbf{y}}_{1},\cdots ,\bar{\mathbf{y}}_{k}, \cdots, \bar{\mathbf{y}}_{\!K} \right \}$, for each target segment $\bar{\mathbf{y}}_{k}\!=\!(y_{{b}_{k}^{y}}, \cdots, y_{e_{k}^{y}})$ where ${b}_{k}^{y}$ and $e_{k}^{y}$ are its beginning and end target positions, we can get the corresponding source segment as $\bar{\mathbf{x}}_{k}=(x_{{b}_{k}^{x}}, \cdots, x_{e_{k}^{x}})$ where
\begin{align}
{b}_k^x=\left\{\begin{matrix}
1 & k\!=\!1 \\
d_{{e}_{k-1}^y}+1 & \mathrm{otherwise}
\end{matrix}\right. \label{eq-bkx} 
\end{align}
and 
\begin{gather}
{e}_k^x=d_{{b}_{k}^y}.
\end{gather}
Thus the sentence pairs $\left<\mathbf{x},\: \mathbf{y}\right>$ can be segmented into the sequence of segment pairs as $\left<\bar{\mathbf{x}}_{1},\: \bar{\mathbf{y}}_{1}\right>\mid \cdots \mid \left<\bar{\mathbf{x}}_{k},\: \bar{\mathbf{y}}_{k}\right>$.
By replacing the source words with READ actions and target words with WRITE actions, we can get the action segment pairs. Then, the read/write path is formed by concatenating all the action segment pairs, where the length of the read/write path is equal to the total number of source words and target words. 

For the example in Figure \ref{attn}(a), the sequence of source positions corresponding to WRITE actions for the whole target sentence is $2,2,2,3,3,5$, with the corresponding read/write path $\mathrm{RRWWWRWWRRW}$. Then, we can get the sequence of segment pairs as $\left<x_{1}\,x_{2},\: y_{1}\,y_{2}\,y_{3}\right> \mid \left<x_{3}, \:y_{4}\,y_{5}\right> \mid \left<x_{4}\,x_{5},\:y_{6}\right>$, and thereby the sequence of action segment pairs as $\left<\mathrm{R}\,\mathrm{R}, \mathrm{W}\,\mathrm{W}\,\mathrm{W}\right> \mid \left<\mathrm{R}, \mathrm{W}\,\mathrm{W}\right> \mid \left<\mathrm{R}\,\mathrm{R}, \mathrm{W}\right>$ shown in Figure \ref{attn}(b).

\textbf{Transpose} After getting the sequence of segment pairs, the transposed read/write path can be derived from it. As the transposed read/write path is in the form to fit the backward single-path SiMT, the sequence of segment pairs should also be transposed to fit the backward single-path SiMT. According to duality constraints, the sequence of segment pairs is shared by the forward and backward SiMT, so we only need to exchange the source segment and target segment in each segment pair, that is from $\left<\bar{\mathbf{x}}_{k},\:\bar{\mathbf{y}}_{k}\right>$ to $\left<\bar{\mathbf{y}}_{k},\:\bar{\mathbf{x}}_{k}\right>$ where the beginning and end positions of each source/target segment remain the same. Then we can get the corresponding transposed action segment pairs by replacing target words with WRITE actions and source words with READ actions. In this way, we accomplish the transposing of segment pairs. Let's review the example in Figure \ref{attn}(b), and the sequence of transposed action segment pairs is  
$\left<\mathrm{W}\,\mathrm{W}\,\mathrm{W}, \mathrm{R}\,\mathrm{R}\right> \mid \left<\mathrm{W}\,\mathrm{W}, \mathrm{R}\right> \mid \left<\mathrm{W}, \mathrm{R}\,\mathrm{R}\right>$
as shown in Figure \ref{attn}(c).

\textbf{Merge} By merging the transposed action segment pairs, we can get the transposed read/write path. The goal of the transposing process is to get the transposed writing probability matrix $\bm{\gamma}$ to constrain the excepted writing probability matrix for the backward single-path SiMT. According to the definition of the writing probability matrix, only the last column in the sub-matrix covered by each segment pair corresponds to WRITE actions. Formally, for each transposed segment pair $\left<\bar{\mathbf{y}}_{k},\:\bar{\mathbf{x}}_{k}\right>$, the following elements in $\bm{\gamma}$ should have the greatest probability to perform WRITE actions as $\{\gamma_{b_{k}^{x}e_{k}^{y}}, \cdots, \gamma_{e_{k}^{x}e_{k}^{y}}\}$. For the three sub-matrices shown in Figure \ref{attn}(c), only the elements of the last column correspond to WRITE actions as shown in Figure \ref{attn}(d), which are $\{\gamma_{13}, \gamma_{23}, \gamma_{35}, \gamma_{46}, \gamma_{56}\}$. We employ the $0-1$ distribution to set the value of the elements of $\bm{\gamma}$, where the elements corresponding to WRITE actions are set to $1$ and others are set to $0$. This is equivalent to the situation that the selection probability for the Bernoulli distribution (in Eq.(\ref{eq3})) is $1$. 

\subsection{Training}
\label{sec:dualpath}

Assuming the expected writing probability matrix for the forward single-path SiMT is $\bm{\alpha}^F$ and its transposed expected writing probability matrix is $\bm{\gamma}^F$, and similarly in the backward single-path SiMT, the matrices are $\bm{\alpha}^B$ and $\bm{\gamma}^B$, respectively. We reduce the gap between the read/write path with the transposed path of read/write path in another direction by minimizing $L_{2}$ distance between their corresponding expected writing probability matrix as follows:
\begin{align}
    \Omega ^{F}=&\left \|\bm{\alpha}^{F} -\bm{\gamma} ^{B} \right \|_{2} \\
    \Omega ^{B}=&\left \|\bm{\alpha}^{B} -\bm{\gamma} ^{F} \right \|_{2}.
\end{align}

The two $L_{2}$ distances are added to the training loss as a regularization term and the final training loss is
\begin{align}
     \mathcal{L}=\mathcal{L}\!\left (\! \bm{\theta }^{F}\! \right )+\mathcal{L}\!\left (\! \bm{\theta }^{B}\! \right )+\lambda_{dual}(\Omega ^{F}+\Omega ^{B}), \label{eq22}
\end{align}
where $\mathcal{L}\!\left ( \bm{\theta }^{F} \right )$ and $\mathcal{L}\!\left ( \bm{\theta }^{B} \right )$ are the loss function of the forward and backward single-path SiMT model respectively, calculated as Eq.(\ref{eq15}). $\lambda_{dual}$ is a hyperparameter and is set to $\lambda_{dual}=1$ in our experiments.

In the inference time, the forward and backward single-path SiMT models can be used separately,  depending on the required translation direction.

\section{Related Work}

\textbf{Dual learning} is widely used in dual tasks, especially machine translation. For both unsupervised \cite{NIPS2016_5b69b9cb,artetxe-etal-2019-effective,sestorain2019zeroshot} and supervised NMT \cite{pmlr-v70-xia17a,WangXZBQLL18}, dual learning can provide additional constraints by exploiting the dual correlation. Unlike most previous dual learning work on NMT, which use the reconstruction between source and target sequences, we focus on SiMT-specific read/write path and explorer its intrinsic properties.

\textbf{SiMT policy} falls into two categories: fixed and adaptive. For fixed policy, the read/write path is defined by rules and fixed during translating. \citet {dalvi-etal-2018-incremental} proposed STATIC-RW, , which alternately read and write $RW$ words after reading $S$ words. \citet {ma-etal-2019-stacl} proposed wait-k policy, which always generates target $k$ tokens lagging behind the source input. \citet{multipath} enhanced wait-k policy by sampling different $k$ during training. \citet{han-etal-2020-end} applied meta-learning in wait-k. \citet{future-guided} proposed future-guided training to apply a full-sentence MT model to guide wait-k policy. \citet{zhang-feng-2021-icts} proposed a char-level wait-k policy. \citet{zhang-feng-2021-universal} proposed a universal SiMT with mixture-of-experts wait-k policy to perform SiMT under arbitrary latency levels.

For adaptive policy, the read/write path is learned and adaptive to the current context. Early adaptive policies used segmented translation \cite{bangalore-etal-2012-real,Cho2016,siahbani-etal-2018-simultaneous}. \citet{gu-etal-2017-learning} trained an agent with reinforcement learning. \citet {Alinejad2019} added a predict operation based on \citet{gu-etal-2017-learning}. \citet{Zheng2019b} trained an agent with golden READ/WRITE actions generated by rules. \citet {Zheng2019a} added a ``delay'' token to read source words. \citet {Arivazhagan2019} proposed MILk, which applied monotonic attention and used a Bernoulli variable to determine writing. \citet {Ma2019a} proposed MMA, which is the implementation of MILk on the Transformer and achieved the current state-of-the-art SiMT performance. \citet{zhang-etal-2020-learning-adaptive} proposed a adaptive segmentation policy. \citet{wilken-etal-2020-neural} used the external ground-truth alignments to train the policy. \citet{liu-etal-2021-cross} proposed cross-attention augmented transducer. \citet{alinejad-etal-2021-translation} introduced a full-sentence model to generate a ground-truth action sequence. \citet{miao-etal-2021-generative} proposed a generative SiMT policy.

The previous methods often lack the internal supervision on read/write path. Some works use external information such as alignment or generated rule-based sequences to guide the read/write path \cite{Zheng2019b,zhang-etal-2020-learning-adaptive,wilken-etal-2020-neural,alinejad-etal-2021-translation}. However, these methods rely too much on heuristic rules, and thus their performance is not comparable to jointly optimizing read/write path and translation. Our method internally explorers the duality between the read/write paths in two directions, and accordingly uses the duality to constrain the read/write paths, thereby obtaining better SiMT performance.

\section{Experiments}
\label{experiments}

\subsection{Datasets}
We evaluated our method on four translation directions of the following two public datasets.

\textbf{IWSLT15\footnote{\url{nlp.stanford.edu/projects/nmt/}} English$\leftrightarrow  $Vietnamese (En$\leftrightarrow $Vi)} (133K pairs) \cite{iwslt2015} We use TED tst2012 (1553 pairs) as validation set and TED tst2013 (1268 pairs) as test set. Following \citet{LinearTime} and \citet{Ma2019a}, we replace tokens that the frequency less than 5 by $\left \langle unk \right \rangle$. After replacement, the vocabulary sizes are 17K and 7.7K for English and Vietnamese, respectively.

\textbf{WMT15\footnote{\url{www.statmt.org/wmt15/}} German$\leftrightarrow $English (De$\leftrightarrow $En)} (4.5M pairs) Following \citet{Ma2019a}, we use newstest2013 (3000 pairs) as validation set and newstest2015 (2169 pairs) as test set. BPE \cite{sennrich-etal-2016-neural} is applied with 32K merge operations and the vocabulary is shared across languages.

\begin{figure*}[t]
\centering
\subfigure[En$\rightarrow $Vi]{
\includegraphics[width=1.45in]{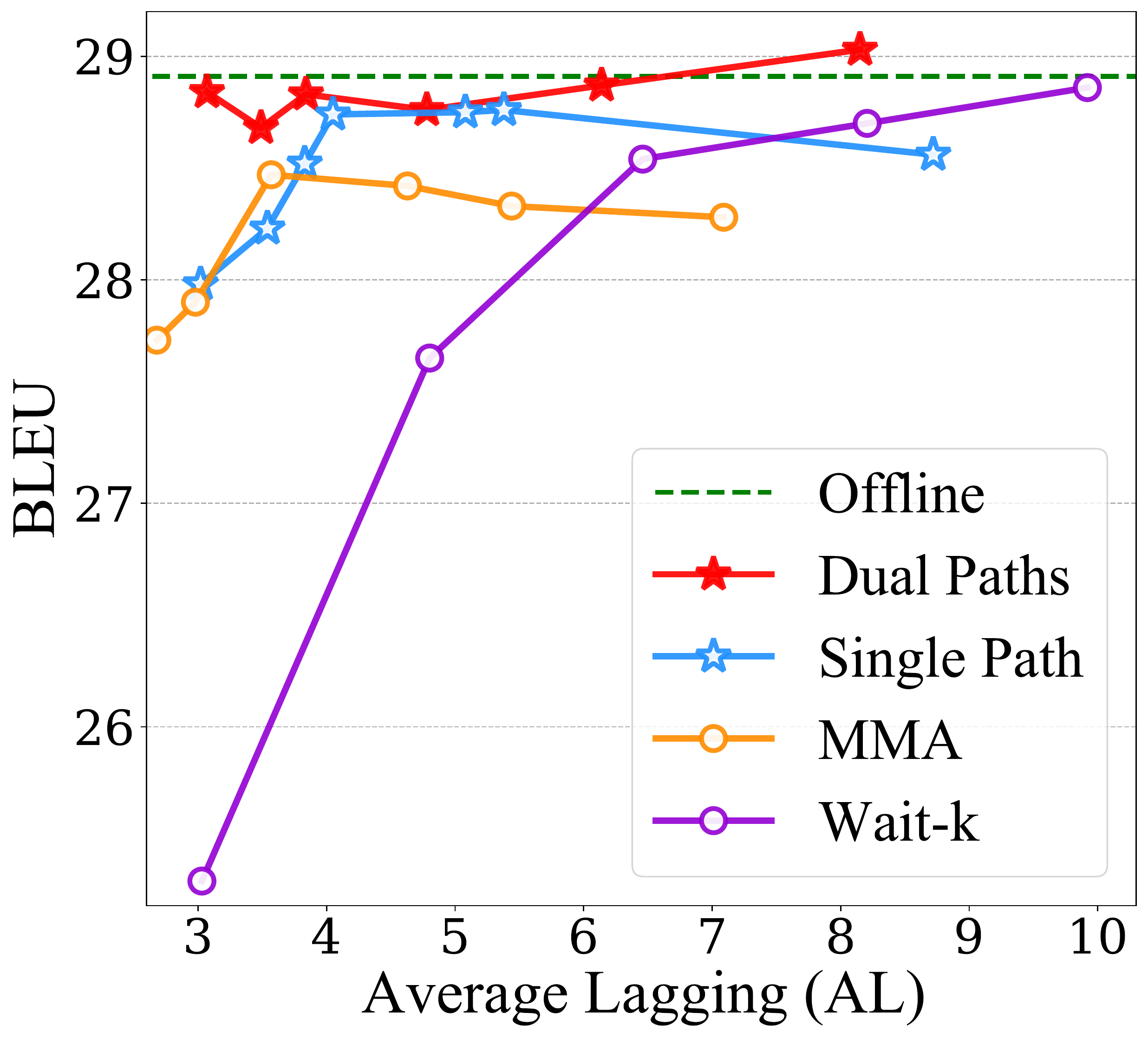}
}
\subfigure[Vi$\rightarrow $En]{
\includegraphics[width=1.45in]{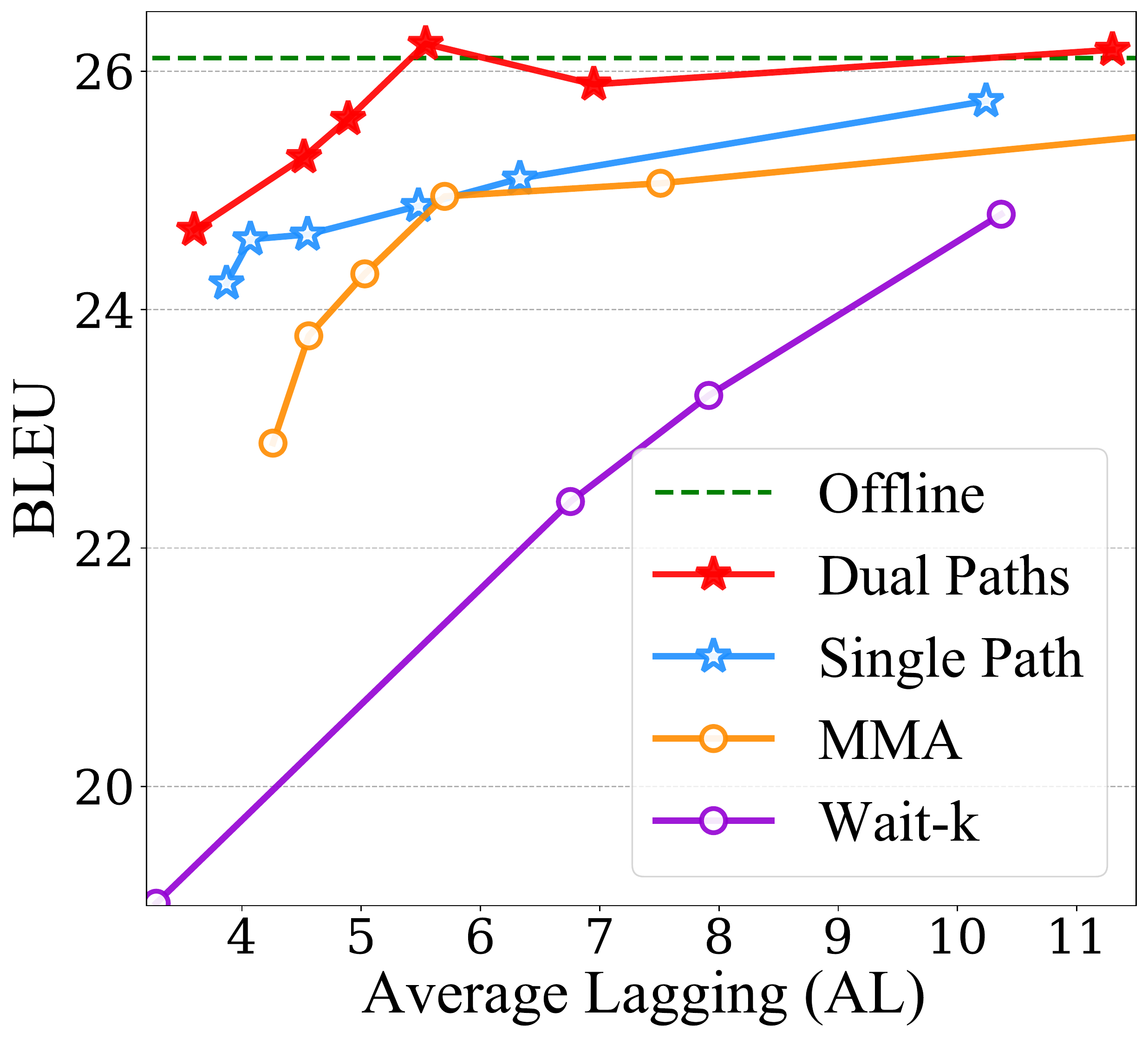}
}
\subfigure[De$\rightarrow $En]{
\includegraphics[width=1.45in]{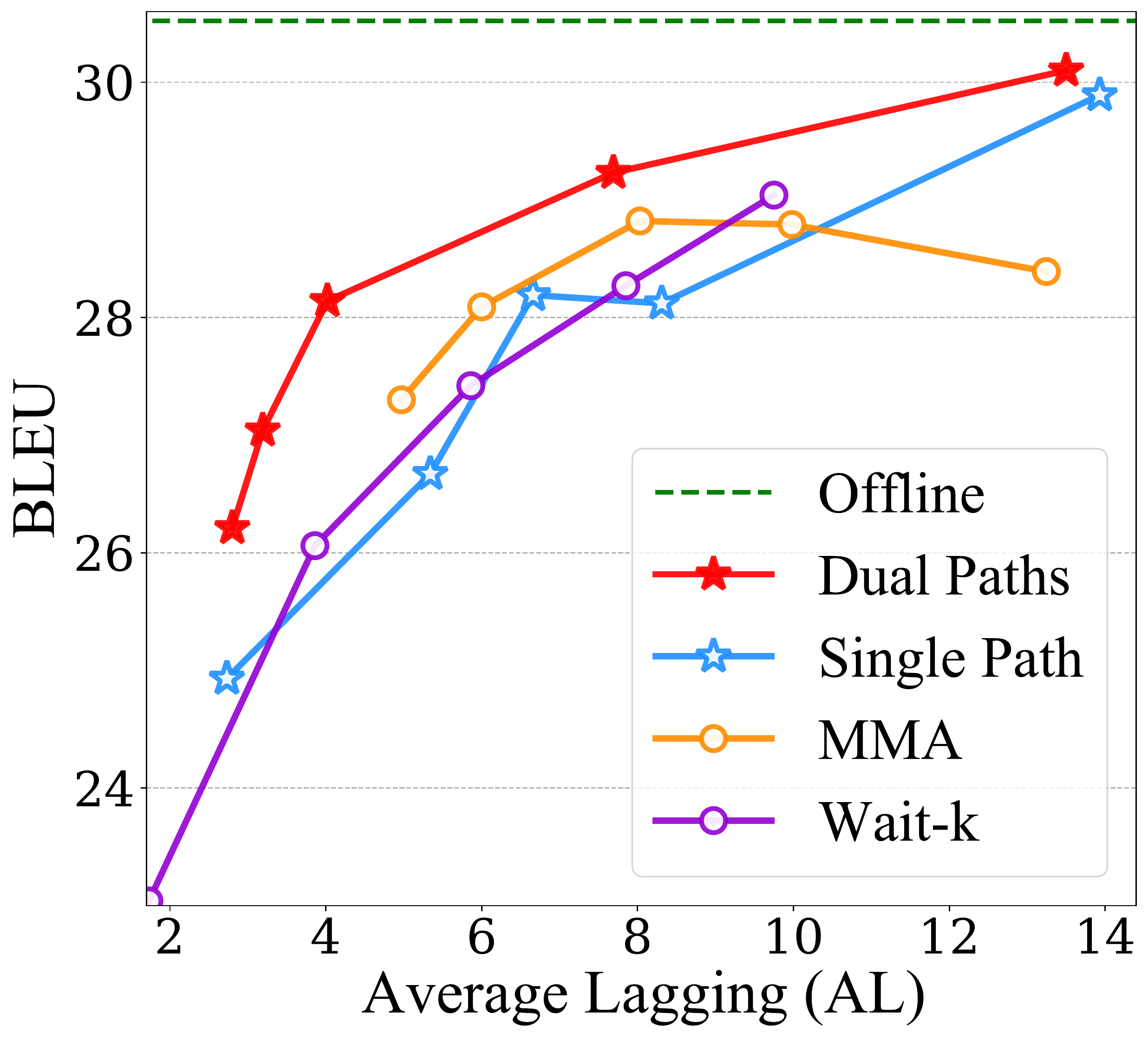}
}
\subfigure[En$\rightarrow $De]{
\includegraphics[width=1.45in]{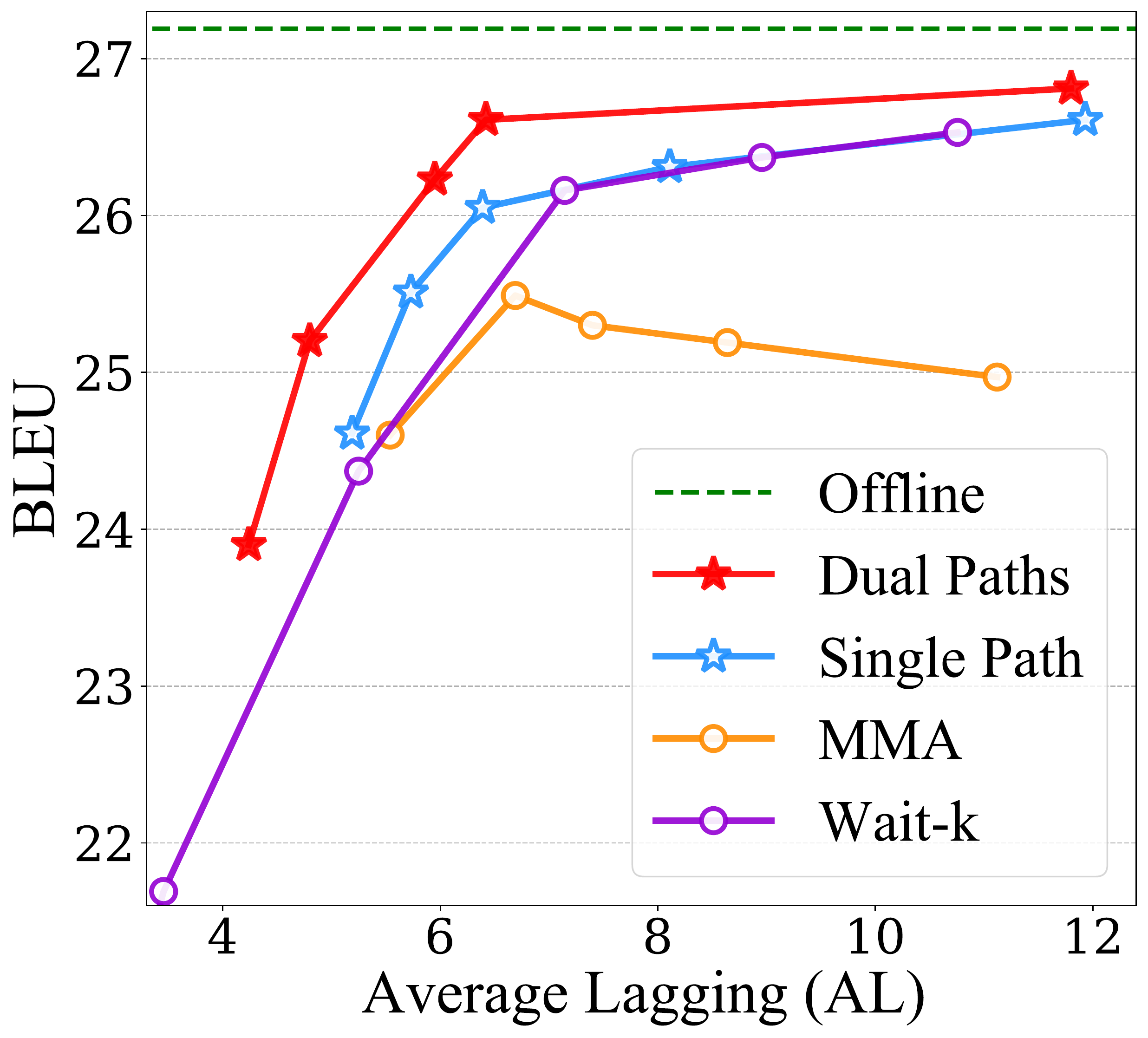}
}

\caption{Translation quality (BLEU) against latency (AL) on the En$\leftrightarrow$Vi and De$\leftrightarrow$En. We show the results of Dual Paths, Single Path, MMA (the current SOTA adaptive policy), Wait-k and Offline model.}
\label{main}
\end{figure*}

\subsection{System Setting}
We conducted experiments on following systems.

{\bf {Offline}} Conventional Transformer \cite{NIPS2017_7181} model for full-sentence translation.

{\bf {Wait-k}} Wait-k policy, the widely used fixed policy \citet{ma-etal-2019-stacl}, which first reads $k$ source tokens and then writes a target word and reads a word alternately.

{\bf {MMA}\footnote{\url{github.com/pytorch/fairseq/tree/master/examples/simultaneous_translation}}} Monotonic multi-head attention (MMA) proposed by \cite{Ma2019a}, the state-of-the-art adaptive policy for SiMT, which applies monotonic attention on each head in Transformer.

{\bf {Single Path}} SiMT model of one translation direction based on monotonic multi-head attention. To avoiding outlier heads\footnote{Since MMA requires all heads in decoder layers to independently decide READ/WRITE action and starts translating only when all heads select WRITE action, some outlier heads that perform too many READ actions will result in higher latency. \citet{Ma2019a} try to control this phenomenon by adding some loss functions, but it still cannot avoid some outlier heads waiting for too many words, which seriously affects the impair the necessity between the READ/WRITE actions in read/write path \cite{Ma2019a,zaidi2021infusing}.} that are harmful for the read/write path, we slightly modified MMA for more stable performance. We no longer let the heads in all decoder layers independently determine the READ/WRITE action, but share the READ/WRITE action between the decoder layers.

{\bf {Dual Paths}} Dual-path SiMT described in Sec.\ref{sec:method}.

The implementations of all systems are adapted from Fairseq Library \cite{ott-etal-2019-fairseq}, based on Transformer \cite{NIPS2017_7181}, where we apply Transformer-Small (4 heads) for En$\leftrightarrow $Vi, and Transformer-Base (8 heads) for De$\leftrightarrow $En. For `Dual Paths', the forward and backward models are used to complete the SiMT on two translation directions at the same time. To perform SiMT under different latency, we set various lagging numbers\footnote{For both En$\leftrightarrow$Vi and De$\leftrightarrow$En: $k=1,3,5,7,9$} $k$ for `Wait-k', and set various latency weights\footnote{For En$\leftrightarrow$Vi: $\lambda=0.01,0.05,0.1,0.2,0.3,0.4$}\footnote{For De$\leftrightarrow$En: $\lambda=0.1,0.2,0.25,0.3,0.4$} $\lambda$ for `MMA', `Single Path' and `Dual Paths'.

We evaluate these systems with BLEU \cite{papineni-etal-2002-bleu} for translation quality and Average Lagging (AL) \cite{ma-etal-2019-stacl} for latency. Average lagging evaluates the number of words lagging behind the ideal policy. Given read/write path $g_{i}$, AL is calculated as
\begin{gather}
    \mathrm{AL}=\frac{1}{\tau }\sum_{i=1}^{\tau}g_{i}-\frac{i-1}{\left | \mathbf{y} \right |/\left | \mathbf{x} \right |},
\end{gather}
where $\tau =\underset{i}{\mathrm{argmax}}\left ( g_{i}= \left | \mathbf{x} \right |\right )$, and $\left | \mathbf{x} \right |$ and $\left | \mathbf{y} \right |$ are the length of the source sentence and target sentence respectively. The results with more latency metrics are shown in Appendix \ref{app:res}.

\subsection{Main Results}
\label{sec:main}

Figure \ref{main} shows the comparison between our method and the previous methods on 4 translation directions. `Dual Paths' outperforms the previous methods under all latency, and more importantly, the proposed duality constraints can improve the SiMT performance on both source-to-target and target-to-source directions concurrently.

\begin{table}[]
\centering
\begin{tabular}{L{3cm}|C{1.2cm}C{1.2cm}} \hlinew{0.7pt}
          & \textbf{AL}   & \textbf{BLEU}  \\\hline
Dual Paths & 7.69 & 29.23 \\ \hline
$\;\;$-w/o Segment & 7.61 & 27.24 \\
$\;\;$-w/o $\Omega ^{B}$     & 8.57 & 28.66 \\
$\;\;$-w/o $\Omega ^{F}, \Omega ^{B}$     & 8.31 & 28.12 \\\hlinew{0.7pt}
\end{tabular}
\caption{Ablation study with $\lambda=0.2$. `w/o Segment': remove the segment operation in transposing process of read/write path, and directly perform transposition. `w/o $\Omega ^{B}$': remove $\Omega ^{B}$ in Eq.(\ref{eq22}), only constrain forward model. `w/o $\Omega ^{F}, \Omega ^{B}$': remove the duality constraints between read/write paths.}
\label{ablation}
\end{table}

Compared to `Wait-k', our method has significant improvement, especially under low latency, since the read/write path in `Wait-k' is fixed and cannot be adjusted. Compared to `MMA', the state-of-the-art adaptive policy, our `Single Path' achieves comparable performance and is more stable under high latency. `MMA' allows each head of each layer to independently predict a read/write path, where some outlier heads will affect the overall performance, resulting in a decline in translation quality under high latency \cite{Ma2019a}. Our method applies a common read/write path instead of the heads in each layer to predict read/write, thereby reducing the possibility of outlier heads. Based on `Single Path', `Dual Paths' further improves the SiMT performance by modeling the duality constraints between read/write paths, especially under low latency. Besides, our method improves the SiMT performance even close to the full-sentence MT on En$\leftrightarrow $Vi, which shows that the more precise read/write path is the key to SiMT performance. Additionally, under the same latency weight $\lambda$, our method tends to have lower latency than `MMA' on De$\leftrightarrow $En. The `Single Path' reduces the unnecessary latency caused by outlier heads, and the duality constraints further improve the necessity of reading source content, thereby achieving lower latency.

\section{Analysis}
We conducted extensive analyses to understand the specific improvements of our method. Unless otherwise specified, all results are reported on De$\rightarrow$En.

\subsection{Ablation Study}

We conducted ablation studies on the duality constraints, where we use direct transposition to replace transposing process of read/write path, only constrain the forward single-path model or remove the duality constraints. As shown in Table \ref{ablation}, the proposed method of transposing the read/write path is critical to translation quality, showing the importance of the segment operation. Besides, mutual constraining between forward and backward single-path model is more conducive to SiMT performance than only constraining one of them or removing constraints.

\subsection{Evaluation of Read/Write Path}
\label{sec:eval_path}

The read/write path needs to ensure sufficient content for translation and meanwhile avoid unnecessary latency, where the aligned source position\footnote{For many-to-one alignment from source to target, we choose the furthest source word. For the target words with no alignment, we ignore them.} is always considered as the oracle position to perform WRITE in previous work \cite{wilken-etal-2020-neural, arthur-etal-2021-learning}. Therefore, we propose two metrics $A^{\!Suf}$ and $A^{\!Nec}$ to measure the \emph{sufficiency} and \emph{necessity} between the READ/WRITE actions in a path via alignments. We denote the ground-truth aligned source position of the $i^{th}$ target word as $a_{i}$, and the read/write path is represented by $g_{i}$, which is the number of source words read in when writing the $i^{th}$ target word. For sufficiency, $A^{\!Suf}$ is used to evaluate whether the aligned source word is read before writing the target word, calculated as
\begin{gather}
    A^{\!Suf}=\frac{1}{I}\sum_{i=1}^{I}\mathbbm{1}_{a_{i}\leq g_{i}},
\end{gather}
where $\mathbbm{1}_{a_{i}\leq g_{i}}$ counts the number of $a_{i}\leq g_{i}$, and $I$ is the target length. For necessity, $A^{\!Nec}$ is used to measure the distance between the output position $g_{i}$ and the aligned source position $a_{i}$, calculated as
\begin{gather}
    A^{\!Nec}=\frac{1}{\left | a_{i}\leq g_{i} \right |}\sum_{i, a_{i}\leq g_{i}}\frac{a_{i}}{g_{i}},
\end{gather}
where the best case is $A^{\!Nec}=1$ for $g_{i}=a_{i}$, performing WRITE just at the aligned position and there is no unnecessary waiting. The more detailed description please refers to Appendix \ref{sec:appendix1}.

\begin{figure}[t]
\centering
\subfigure[Sufficiency of read/write path $A^{\!Suf}$ $\uparrow$.]{
\includegraphics[width=2.6in]{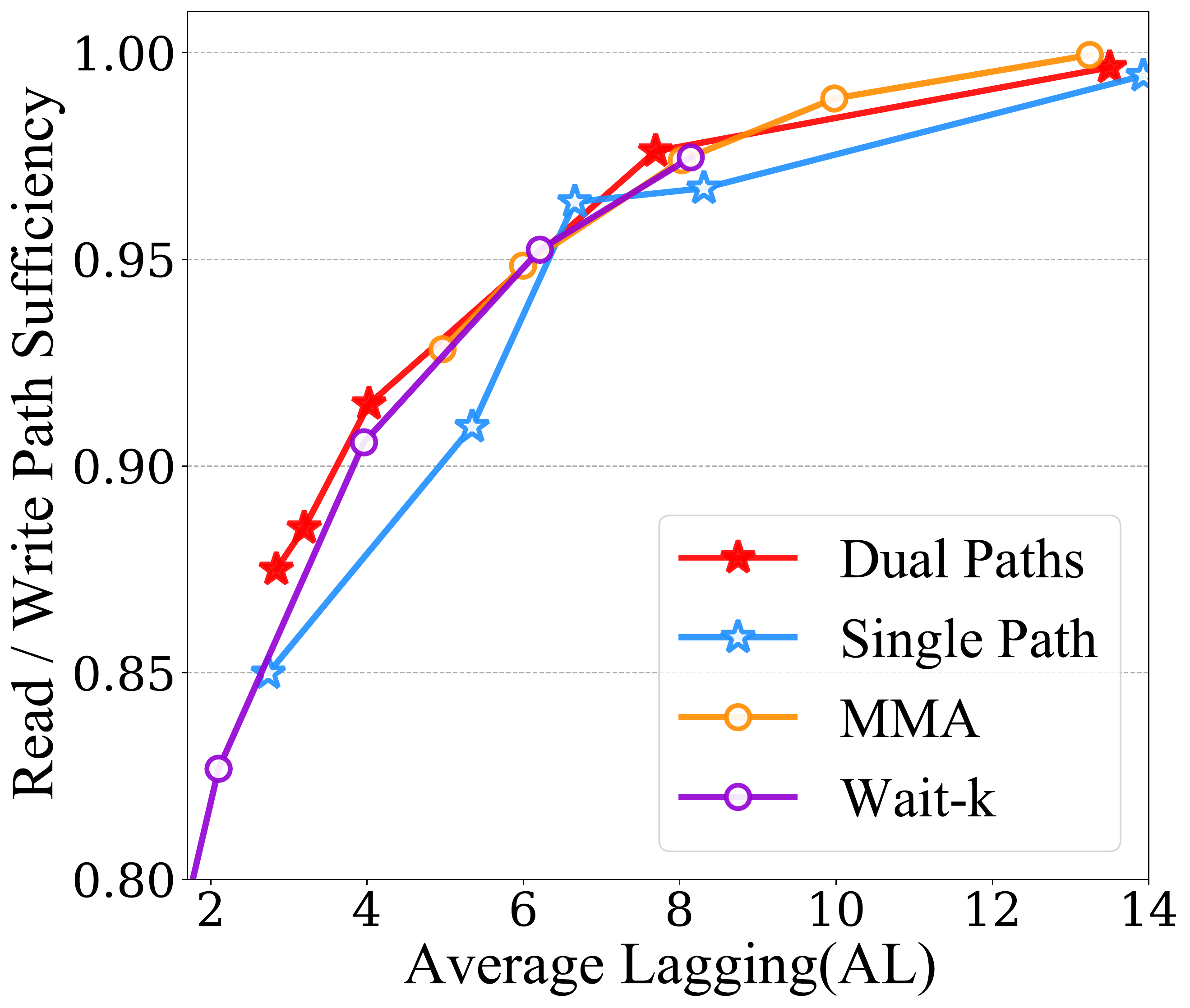}
}
\subfigure[Necessity of read/write path $A^{\!Nec}$ $\uparrow$.]{
\includegraphics[width=2.6in]{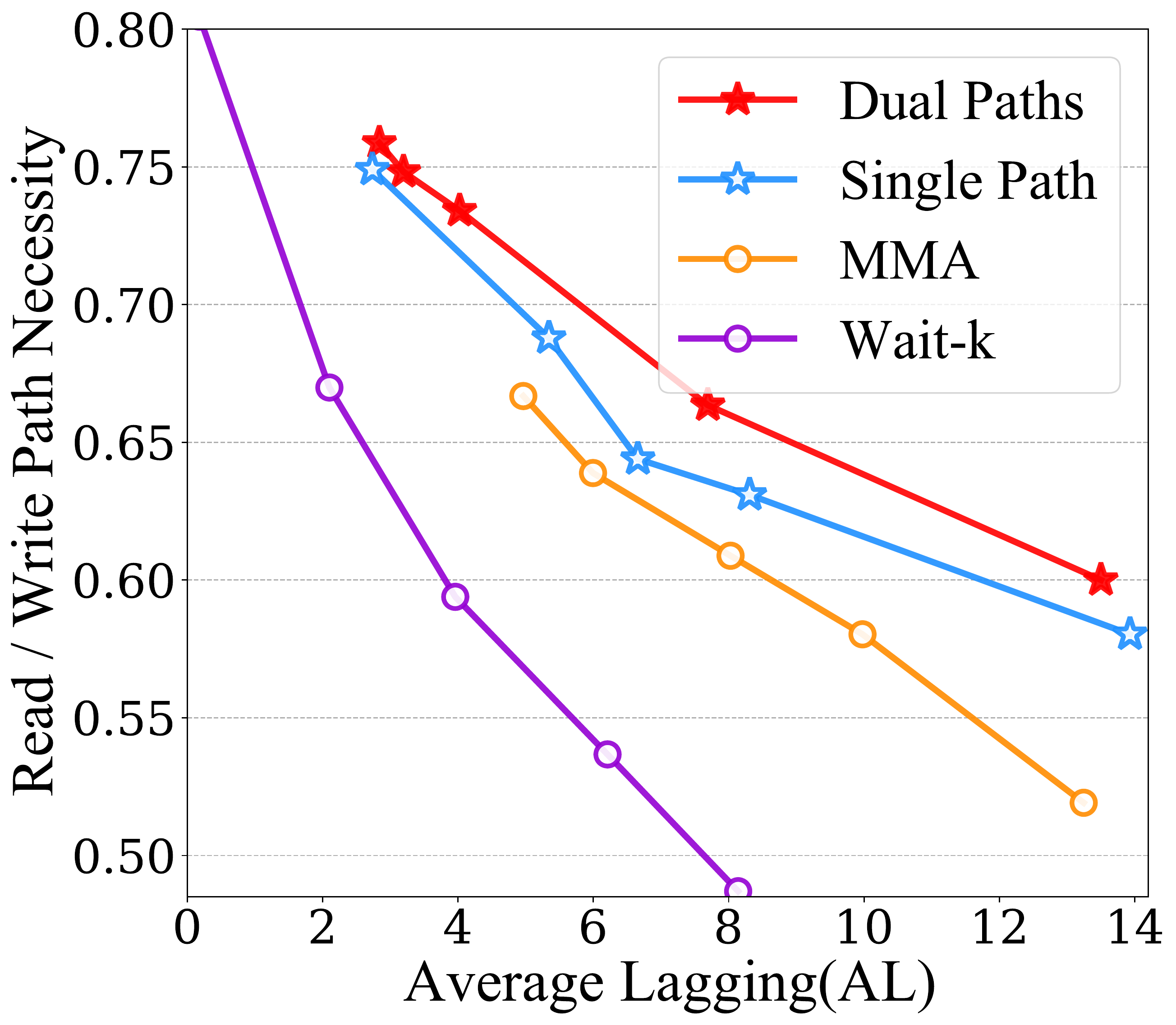}
}

\caption{Sufficiency evaluation and necessity evaluation of the read/write path.}
\label{rwpath}
\end{figure}

As shown in Figure \ref{rwpath}, we evaluate the $A^{\!Suf}$ and $A^{\!Nec}$ of read/write path on RWTH De$\rightarrow$En alignment dataset \footnote{\url{https://www-i6.informatik.rwth-aachen.de/goldAlignment/}},  whose reference alignments are manually annotated by experts. The read/write paths of all methods perform similarly in sufficiency evaluation and our method performs slightly better at low latency. Except that the fixed policy `Wait-k' may be forced to start translating before reading the aligned source word under the lower latency, `MMA' and our method can almost cover more than 85\% of the aligned source word when starting translating. In the necessity evaluation, our method surpasses `Wait-k' and `MMA', and starts translation much closer to the aligned source word, which shows that duality constraints make read/write path more precise, avoiding some unnecessary waiting. Note that while avoiding unnecessary waiting, our method also improves the translation quality (see Figure \ref{main}) under the same latency, which further shows the importance of a proper read/write path for SiMT performance.

\subsection{Effect of Duality Constraints}
\label{sec:duality}

\begin{table}[]
\centering
\begin{tabular}{L{1.2cm}|C{1.4cm}C{1.4cm}C{1.4cm}} \hlinew{0.7pt}
\multirow{2}{*}{\textbf{Latency}} & \multicolumn{3}{c}{\begin{tabular}[c]{@{}c@{}}\textbf{Duality of Read/Write Path (IoU)}\\ \textbf{between De$\rightarrow $En and En$\rightarrow $De}\end{tabular}} \\ \cline{2-4}
                     & \multicolumn{1}{c}{MMA} & \multicolumn{1}{c}{Single Path} & \multicolumn{1}{c}{Dual Path} \\\hline
\textbf{High}                                         & 0.4755                  & 0.5328                      & 0.6346                        \\
\textbf{Middle}                                        & 0.5132                  & 0.5898                      & 0.6962                        \\
\textbf{Low}                                         & 0.6046                  & 0.7169                      & 0.7466                        \\\hlinew{0.7pt}             
\end{tabular}
\caption{Duality of read/write path (IoU score) between De$\rightarrow$En and En$\rightarrow$De.}
\label{sam}
\end{table}

To verify that our method improves the duality of two read/write paths, we conduct duality evaluation between source-to-target and target-to-source read/write paths. Specifically, we first express both the original read/write path on target-to-source and the transposed path of source-to-target read/write path in the form of matrices, and then calculate the \emph{Intersection over Union score} (IoU) between the area below them (see Figure \ref{vis}), which is regarded as the duality between the read/write path in the two directions. The higher IoU score indicates that the two paths are more consistent on common segment pairs, i.e., stronger duality. Appendix \ref{sec:appendix2} gives the detailed calculation of IoU score.

The results of duality evaluation are reported in Table \ref{sam}, where our method effectively enhances the duality of source-to-target and target-to-source read/write paths under all latency levels. This shows that with dual-path SiMT, the read/write paths in source-to-target and target-to-source are more in agreement on the sequence of segment pairs between the sentence pair.

\subsection{Dual Read/Write Paths Visualization}
\begin{figure*}[t]
\centering
\includegraphics[width=6.23in]{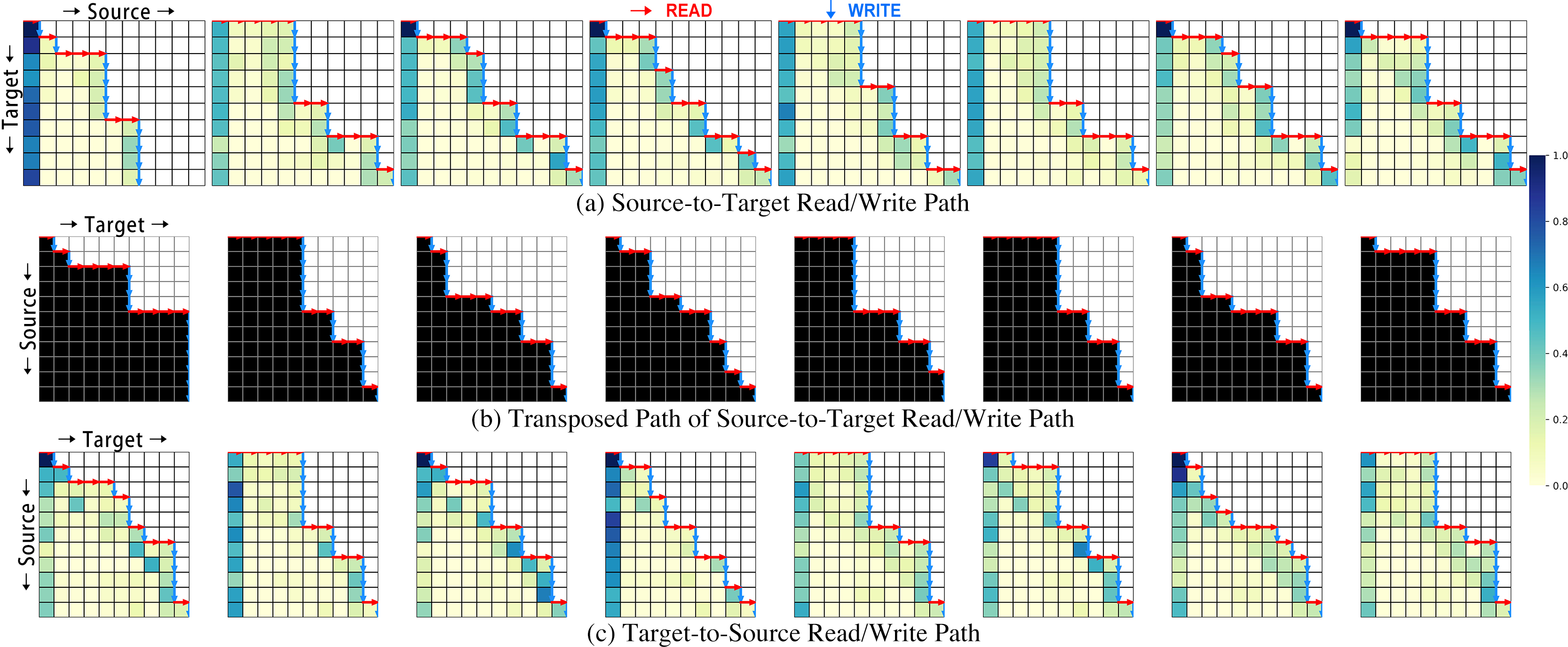}
\caption{Read/write path visualization of a De$\leftrightarrow$En example (De: `\textit{die Lehr@@ er@@ bildung fand in Bam@@ berg statt .}' $\leftrightarrow$ En: `\textit{the teacher training course was in Bam@@ berg .}'). (a) and (c) show the read/write path and attention distribution in two single-path SiMT model, where the shade of the color indicates the attention weight. (b) shows the transposed path of the source-to-target read/write path. `\textcolor{red}{$\rightarrow$}': READ action to wait for a source word, `\textcolor{blue}{$\downarrow$}': WRITE action to generate a target word. Note that 8 sub-figures respectively represent 8 read/write paths, assigned to 8 heads and shared between decoder layers, and the attention is averaged on all decoder layers.}
\label{vis}
\end{figure*}

Figure \ref{vis} shows the read/write path visualization of a De$\leftrightarrow$En example. In `Dual Paths', there is a strong duality between the read/write paths in two translation directions, where the target-to-source read/write path (Figure \ref{vis}c) and the transposed path of the source-to-target read/write path (Figure \ref{vis}b) have a high degree of overlap. In particular, the read/write paths in our method exhibit a clear division on segment pairs.

\subsection{Analysis on Forward/Backward Latency}
\label{sec:lamda}

\begin{table*}[]
\centering
\begin{tabular}{L{2.2cm}|C{0.7cm}C{0.7cm}|C{1.2cm}C{1.2cm}|C{1.2cm}C{1.2cm}} \hlinew{0.7pt}
\multirow{2}{*}{\textbf{Systems}}                                             & \multirow{2}{*}{$\lambda^{F}$} & \multirow{2}{*}{$\lambda^{B}$ } & \multicolumn{2}{c|}{\textbf{De}$\rightarrow$\textbf{En}} & \multicolumn{2}{c}{\textbf{En}$\rightarrow$\textbf{De}} \\ 
                                                                     &                    &                    & \textbf{AL}               & \textbf{BLEU}             & \textbf{AL}               & \textbf{BLEU}              \\ \hline
MMA                                                                  & 0.3                & -                  & 6.00             & 27.29            & -                & -                 \\ \hline
Single Path                                                              & 0.3                & -                  & 5.34             & 26.67            & -                & -                 \\ \hline
\multirow{3}{*}{\begin{tabular}[c]{@{}c@{}}Dual Paths\end{tabular}} & 0.3                & 0.2                & 4.71             & 27.39            & 6.43             & 25.53             \\
                                                                     & 0.3                & 0.3                & 3.19             & 27.04            & 4.80             & 25.20             \\
                                                                     & 0.3                & 0.4                & 3.00             & 27.01            & 3.77             & 23.62            \\\hlinew{0.7pt}
\end{tabular}
\caption{Performance under different settings of latency weight, where $\lambda^{F}$ and $\lambda^{B}$ are the latency weight of the forward and backward single-path SiMT model respectively.}
\label{lambda}
\end{table*}

To analyze the relationship between the forward and backward single-path SiMT model in terms of the latency setting, we set the latency weight ($\lambda$ in Eq.(\ref{eq15})) of the forward and backward single-path SiMT model to different values, denoted as $\lambda^{F}$ and  $\lambda^{B}$ respectively (the greater the latency weight, the lower the model latency). Table \ref{lambda} reports the effect of different settings of $\lambda^{B}$ on the performance of the forward single-path model.

After applying backward model and the duality constraints, our method has a much lower latency and similar translation quality compared with `MMA' and `Single Path'. As the latency of the backward model decreases ($\lambda^{B}$ becomes larger), the latency of the forward model also gradually decreases, which shows that the latency of the forward and backward models are strongly correlated. Overall, regardless of the setting of $\lambda^{F}$ and $\lambda^{B}$, `Dual Paths' obtains a better trade-off between latency and translation quality. Furthermore, we can get a slightly larger or smaller latency by adjusting the combination of $\lambda^{F}$ and $\lambda^{B}$.

\section{Conclusion}
In this paper, we develop the dual-path SiMT to supervise the read/write path by modeling the duality constraints between SiMT in two directions. Experiments and analyses we conducted show that our method outperforms strong baselines under all latency and achieves a high-quality read/write path.

\section*{Acknowledgements}
We thank all the anonymous reviewers for their insightful and valuable comments. This work was supported by National Key R\&D Program of China (NO. 2017YFE0192900).

\bibliography{anthology,custom}
\bibliographystyle{acl_natbib}

\clearpage
\appendix

\section{Evaluation Metrics of Read/Write Path}
\label{sec:appendix1}
In Sec.\ref{sec:eval_path}, we propose two metrics $A^{\!Suf}$ and $A^{\!Nec}$ to measure the \emph{sufficiency} and \emph{necessity} of the read/write path using alignments. Here, we give a more detailed calculation of them.

\begin{figure}[t]
\centering
\includegraphics[width=2.8in]{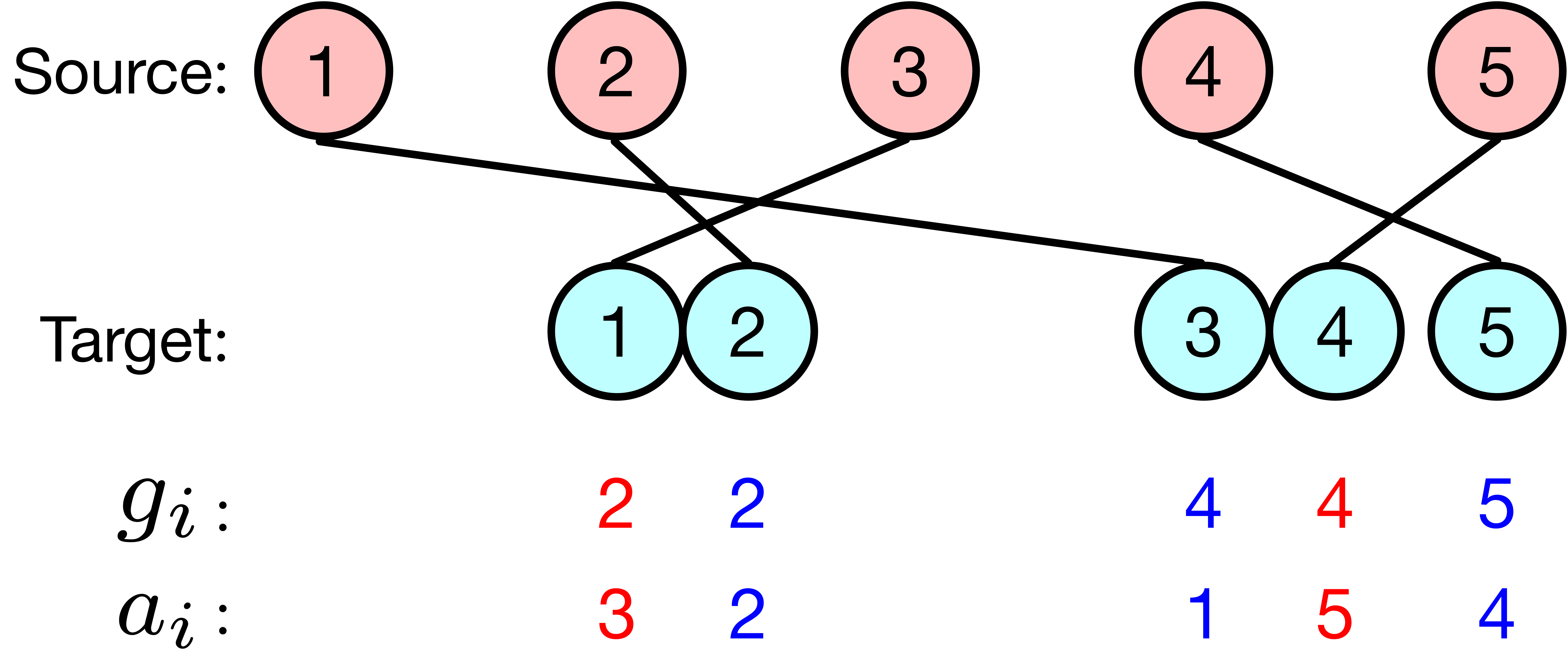}
\caption{Schematic diagram of evaluating the read/write path in terms of sufficiency and necessity. The black line indicates the ground-truth alignments between the target and source. $g_{i}$ is  the number of source words read in when generating the $i^{th}$ target word. $a_{i}$ is the ground-truth aligned source position of the $i^{th}$ target word. $a_{i} > g_{i}$ (numbers colored in red) means that the $i^{th}$ target word is forced to be translated in advance before reading its aligned source word.}
\label{patheval}
\end{figure}

Given the ground-truth alignments, we denote the aligned source position of the $i^{th}$ target word as $a_{i}$. Specifically, for one-to-many alignment from target to source, we choose the furthest source word as it aligned source position. For a read/write path, we denote the number of source words read in when generating the $i^{th}$ target word as $g_{i}$. Figure \ref{patheval} gives an example of the calculation of $a_{i}$ and $g_{i}$.

\textbf{Sufficiency} $A^{\!Suf}$ measures how many aligned source words are read before translating the target word (i.e., $a_{i}\leq g_{i}$), which ensures the faithfulness of the translation, calculated as
\begin{gather}
    A^{\!Suf}=\frac{1}{I}\sum_{i=1}^{I}\mathbbm{1}_{a_{i}\leq g_{i}},
\end{gather}
where $\mathbbm{1}_{a_{i}\leq g_{i}}$ counts the number that $a_{i}\leq g_{i}$. Taking the case in Figure \ref{patheval} as an example, the sufficiency is calculated as $A^{\!Suf}=\frac{1}{5}\times \left (0+1+1+0+1 \right )=\frac{3}{5}$, where the $1^{st}$ and $4^{th}$ target word are translated before read their aligned source word ($a_{i} > g_{i}$).

\textbf{Necessity} $A^{\!Nec}$ measures how far the output position $g_{i}$ is from the aligned position $a_{i}$, where the closer output position to the alignment position indicates that the read/write path outputs earlier, and there is less unnecessary latency. $A^{\!Nec}$ is calculated as
\begin{gather}
    A^{\!Nec}=\frac{1}{\left | a_{i}\leq g_{i} \right |}\sum_{i, a_{i}\leq g_{i}}\frac{a_{i}}{g_{i}},
\end{gather}
Note that $A^{\!Nec}$ only focuses on aligned positions that are read before output position (i.e., $a_{i}\leq g_{i}$). In the case shown in Figure \ref{patheval}, the necessity is calculated as $A^{\!Nec}=\frac{1}{3} \times \left (\frac{2}{2}+\frac{1}{4}+\frac{4}{5} \right )=\frac{41}{60}$, where we only consider the the $2^{th}$, $3^{rd}$ and $5^{th}$ target word.

\section{IoU Score for Duality Evaluation}
\label{sec:appendix2}

\begin{figure}[t]
\centering
\includegraphics[width=2.6in]{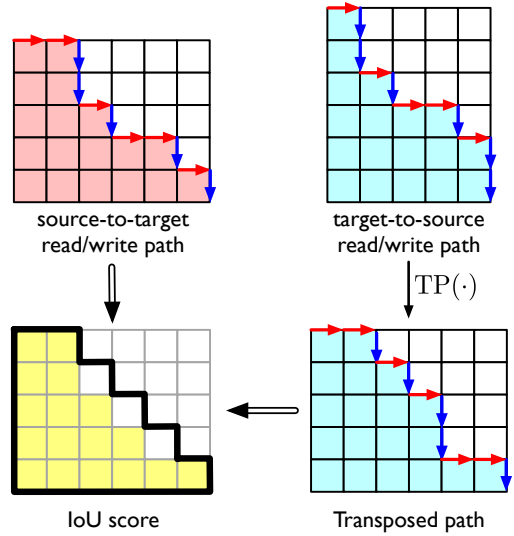}
\caption{Schematic diagram of calculating the Intersection over Union score (IoU) to evaluate the dual degree of source-to-target and target-to-source read/write path. The yellow area represents the union of the areas below two paths, and the area enclosed by the black line represents the intersection.}
\label{iou}
\end{figure}

To verify that our proposed method does make the read/write path of source-to-target and target-to-source more dual, we calculate the \emph{Intersection over Union score} (IoU) to evaluate the duality in Sec.\ref{sec:duality}. Following, we describe the detailed calculation of IoU score.

Figure \ref{iou} gives an example of calculating the IoU score. Given the source-to-target and target-to-source read/write path $\textbf{P}^{s2t}$ and $\textbf{P}^{t2s}$ in the binary matrix form, we first generate the transposed path $\textbf{TP}^{s2t}$ of $\textbf{P}^{t2s}$ with proposed method of transposing the read/write path in Sec.\ref{sec:tp}. Then, we calculate the intersection over union score between binary matrices $\textbf{P}^{s2t}$ and $\textbf{TP}^{s2t}$:
\begin{gather}
    \mathrm{IoU}=\frac{\mathrm{Sum}\left ( \textbf{P}^{s2t}\cap \textbf{TP}^{s2t} \right )}{\mathrm{Sum}\left ( \textbf{P}^{s2t}\cup  \textbf{TP}^{s2t} \right )},
\end{gather}
where the larger IoU score means that the source-to-target and target-to-source read/write path are much more dual. Ideally, the best case is $ \mathrm{IoU}=1$, which means the source-to-target and target-to-source read/write path are exactly in the dual form and reach the agreement on the sequence of segment pairs.

In the calculation of IoU score, for `MMA' and `Single Path', the source-to-target and target-to-source read/write paths come from independent models in the two directions respectively. For `Dual Paths', the source-to-target and target-to-source read/write paths come from the forward and backward single-path SiMT model concurrently.

\section{Hyperparameters}
All systems in our experiments use the same hyperparameters, as shown in Table \ref{Hyper}.

\begin{table*}[]
\centering
\begin{tabular}{L{4.5cm}C{3cm}C{3cm}} \hlinew{1.5pt}
\textbf{Hyperparameter} & \textbf{IWSLT15 En$\leftrightarrow $Vi} & \textbf{WMT15 De$\leftrightarrow $En} \\ \hlinew{1pt} 
encoder layers          & 6                                 & 6                             \\
encoder attention heads & 4                                 & 8                             \\
encoder embed dim       & 512                               & 512                           \\
encoder ffn embed dim   & 1024                              & 1024                          \\
decoder layers          & 6                                 & 6                             \\
decoder attention heads & 4                                 & 8                             \\
decoder embed dim       & 512                               & 512                           \\
decoder ffn embed dim   & 1024                              & 1024                          \\
dropout                 & 0.3                               & 0.3                           \\
optimizer               & adam                              & adam                          \\
adam-$\beta$                  & (0.9, 0.98)                       & (0.9, 0.98)                   \\
clip-norm               & 0                                 & 0                             \\
lr                      & 5e-4                              & 5e-4                          \\
lr scheduler            & inverse sqrt                     & inverse sqrt                 \\
warmup-updates          & 4000                              & 4000                          \\
warmup-init-lr          & 1e-7                              & 1e-7                          \\
weight decay            & 0.0001                            & 0.0001                        \\
label-smoothing         & 0.1                               & 0.1                           \\
max tokens              & 16000                             & 2400$\times$4$\times$8   \\\hlinew{1.5pt}                  
\end{tabular}
\caption{Hyperparameters of our experiments.}
\label{Hyper}
\end{table*}

\section{Numerical Results with More Metrics}
\label{app:res}
We also compare `Dual Paths' and `Single Path' with previous methods on the latency metrics Average Proportion (AP) \cite{Cho2016} and Differentiable Average Lagging (DAL) \cite{Arivazhagan2019}. In this section, we first give the definition of AP and DAL, and then report the expanded results and numerical results of the main experiment (Sec.\ref{sec:main}), using AP, AL, DAL as latency metrics.

\subsection{Latency Metrics}
\textbf{Average Proportion (AP)} \cite{Cho2016} measures the proportion of the area above a read/write path. Given the read/write path $g_{i}$, AP is calculated as
\begin{gather}
    \mathrm{AP}=\frac{1}{\left | \mathbf{x} \right | \left | \mathbf{y} \right |}\sum_{i=1}^{\left | \mathbf{y} \right |} g_{i}.
\end{gather}

\textbf{Differentiable Average Lagging (DAL)} \cite{Arivazhagan2019} is a differentiable version of average lagging, which can be integrated into training. Given the read/write path $g_{i}$, DAL is calculated as
\begin{gather}
g^{'}_{i}=\left\{\begin{matrix}
g_{i} & i=1\\ 
 \mathrm{max}\left (g_{i},g^{'}_{i-1}+ \frac{\left | \mathbf{x} \right |}{\left | \mathbf{y} \right |} \right )& i>1
\end{matrix}\right. , \\
    \mathrm{DAL}=\frac{1}{\left | \mathbf{y} \right | }\sum\limits_{i=1}^{\left | \mathbf{y} \right |}g^{'}_{i}-\frac{i-1}{\left | \mathbf{x} \right |/\left | \mathbf{y} \right |}.
\end{gather}

\subsection{Expand Results}
\begin{figure*}[]
\centering
\subfigure[IWSLT15 En$\rightarrow $Vi, AP]{
\includegraphics[width=3in]{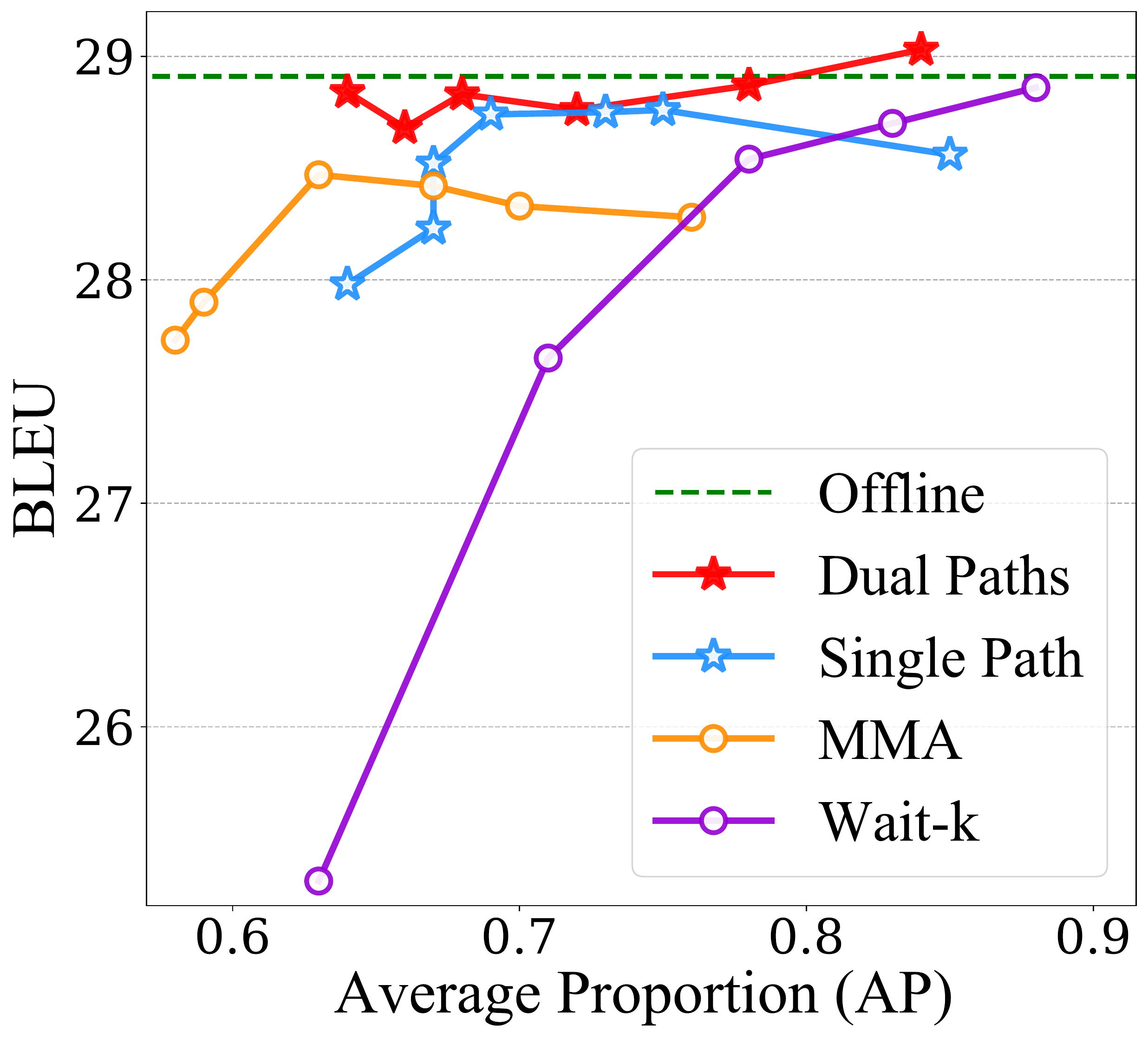}
}
\subfigure[IWSLT15 En$\rightarrow $Vi, DAL]{
\includegraphics[width=3in]{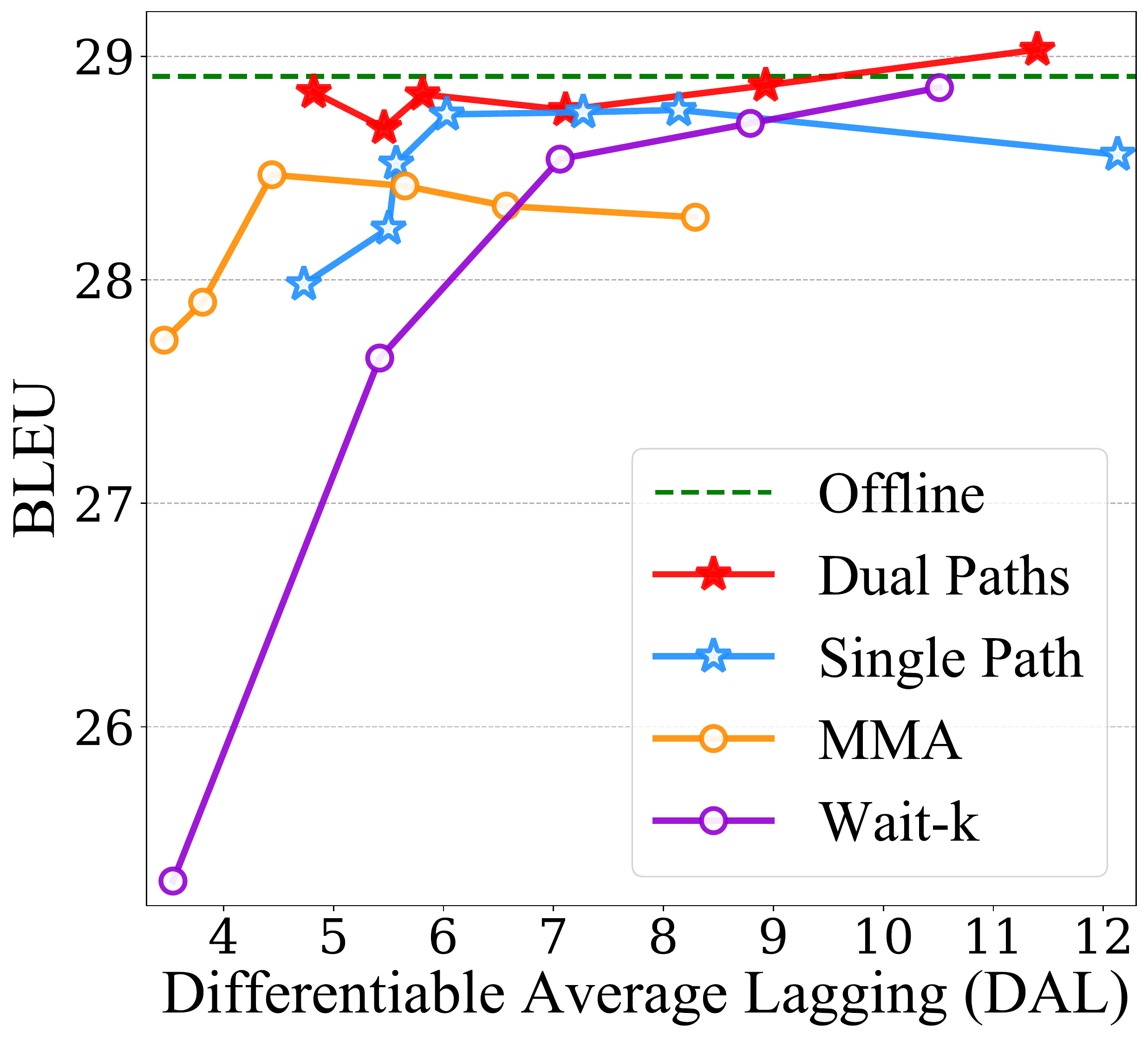}
}
\caption{Results on IWSLT15 En$\rightarrow $Vi, measured with AP and DAL.}
\label{envimain}
\end{figure*}

\begin{figure*}[]
\centering
\subfigure[IWSLT15 Vi$\rightarrow $En, AP]{
\includegraphics[width=3in]{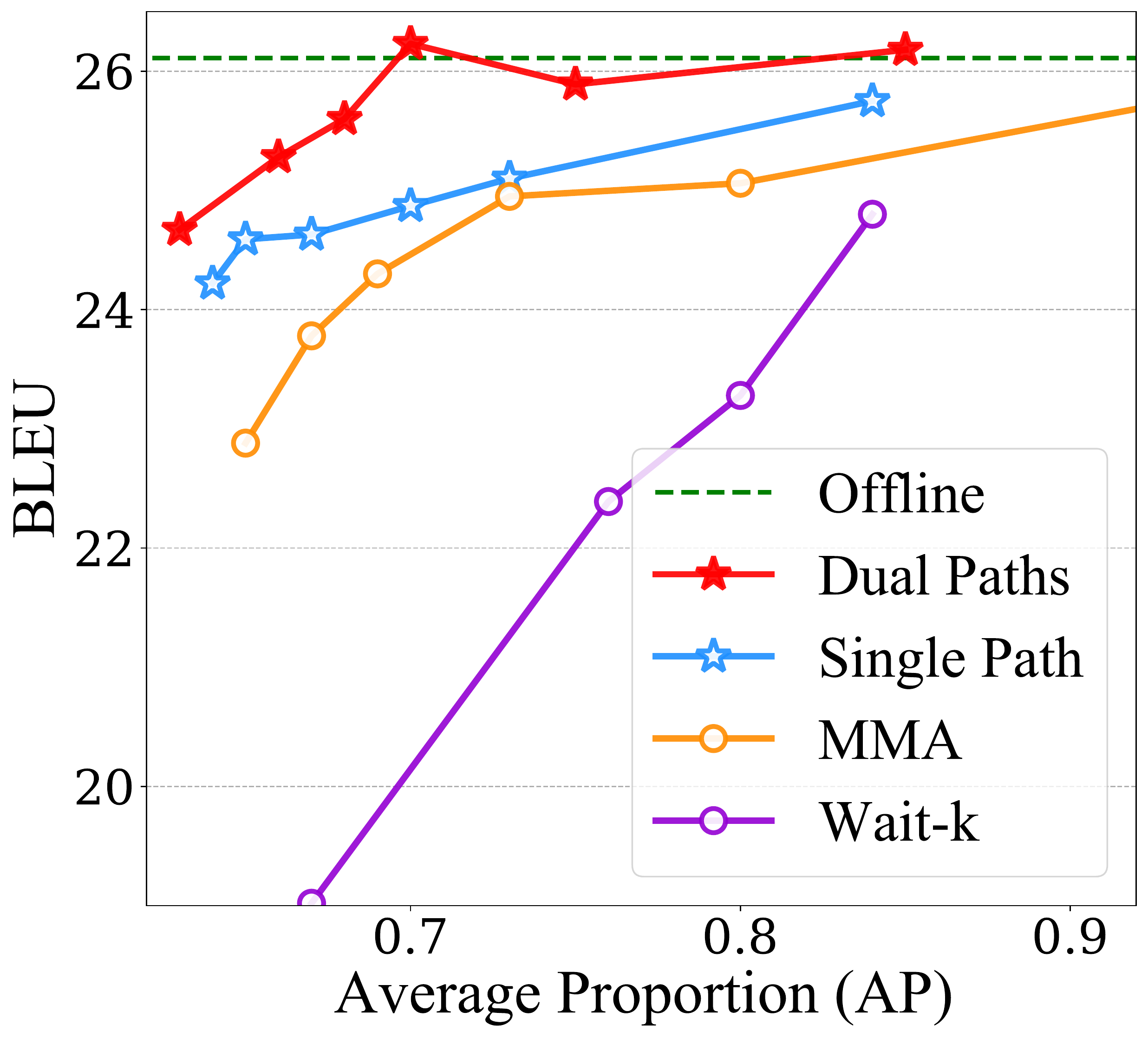}
}
\subfigure[IWSLT15 Vi$\rightarrow $En, DAL]{
\includegraphics[width=3in]{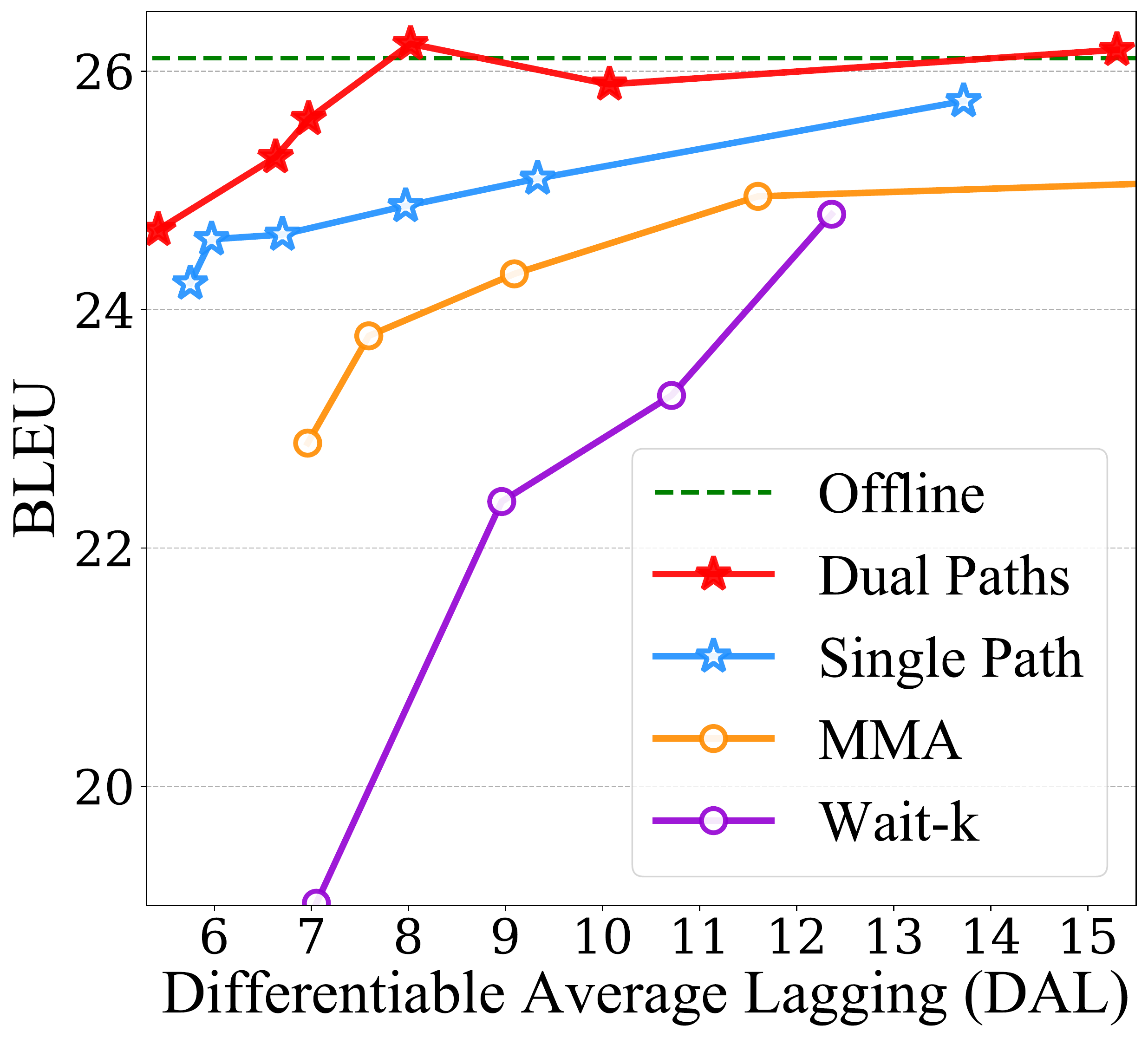}
}
\caption{Results on IWSLT15 Vi$\rightarrow $En, measured with AP and DAL.}
\label{vienmain}
\end{figure*}

\begin{figure*}[]
\centering
\subfigure[WMT15 De$\rightarrow $En, AP]{
\includegraphics[width=3in]{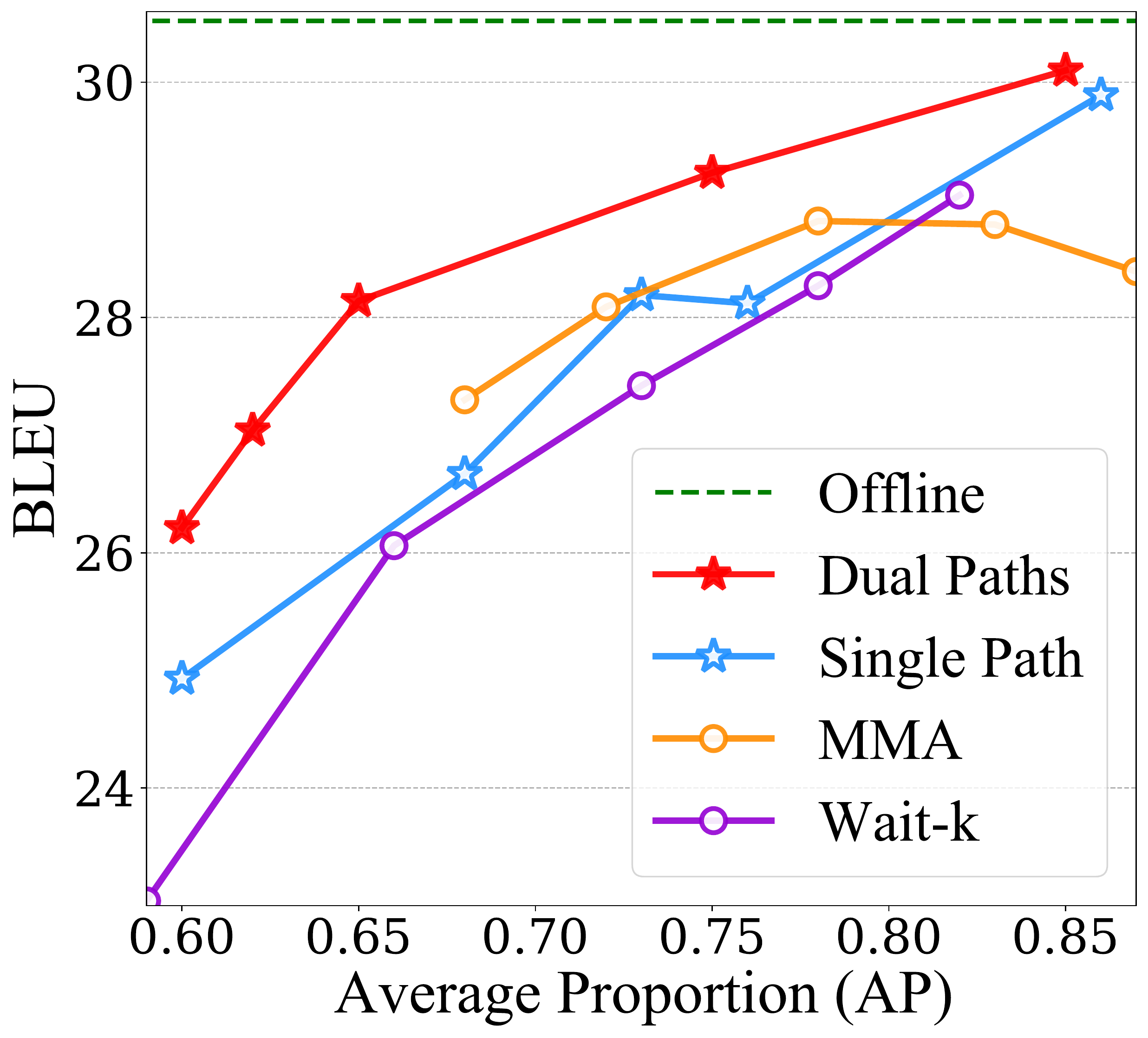}
}
\subfigure[WMT15 De$\rightarrow $En, DAL]{
\includegraphics[width=3in]{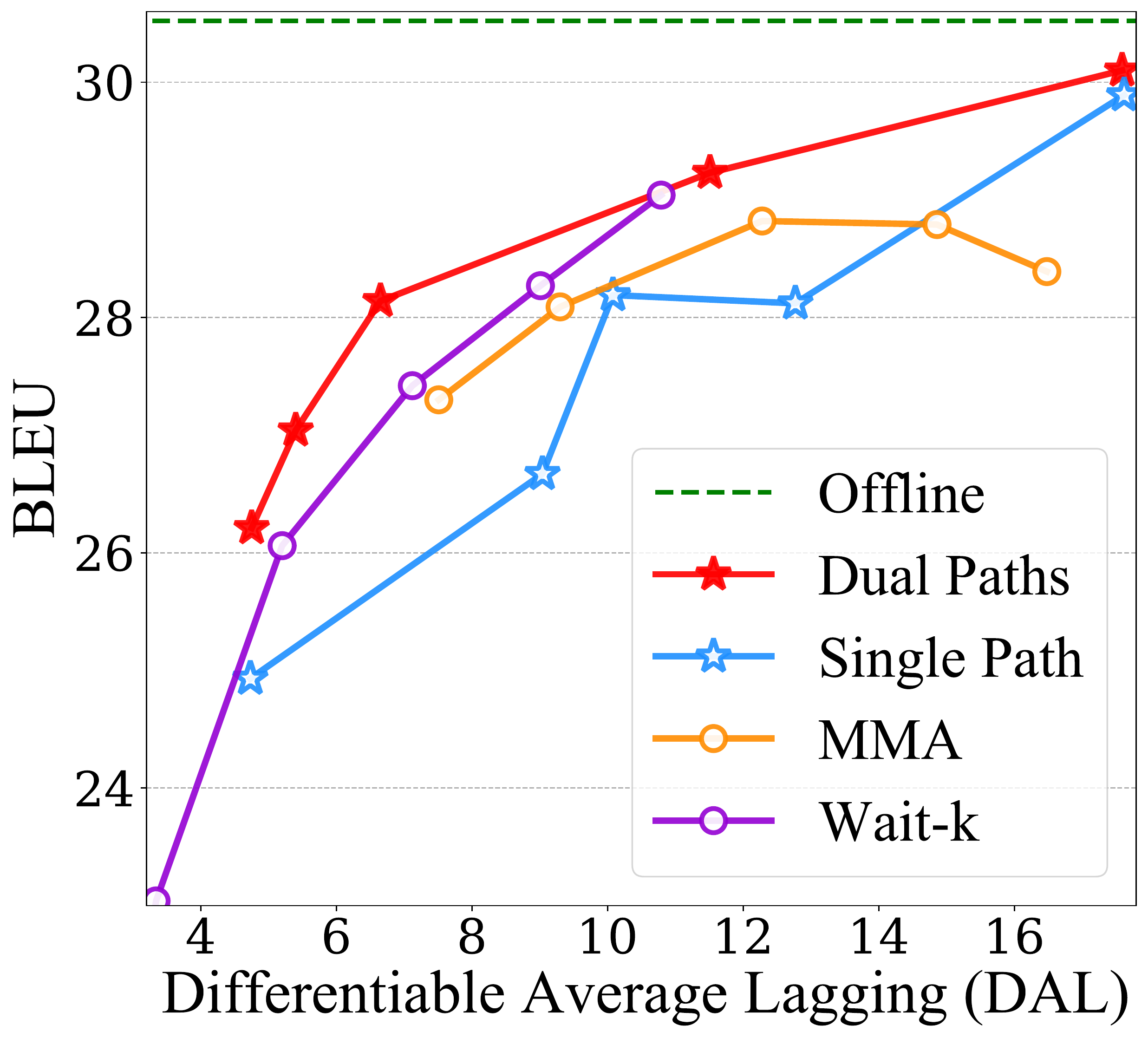}
}

\caption{Results on WMT15 De$\rightarrow $En, measured with AP and DAL.}
\label{deenmain}
\end{figure*}

\begin{figure*}[]
\centering
\subfigure[WMT15 En$\rightarrow $De, AP]{
\includegraphics[width=3in]{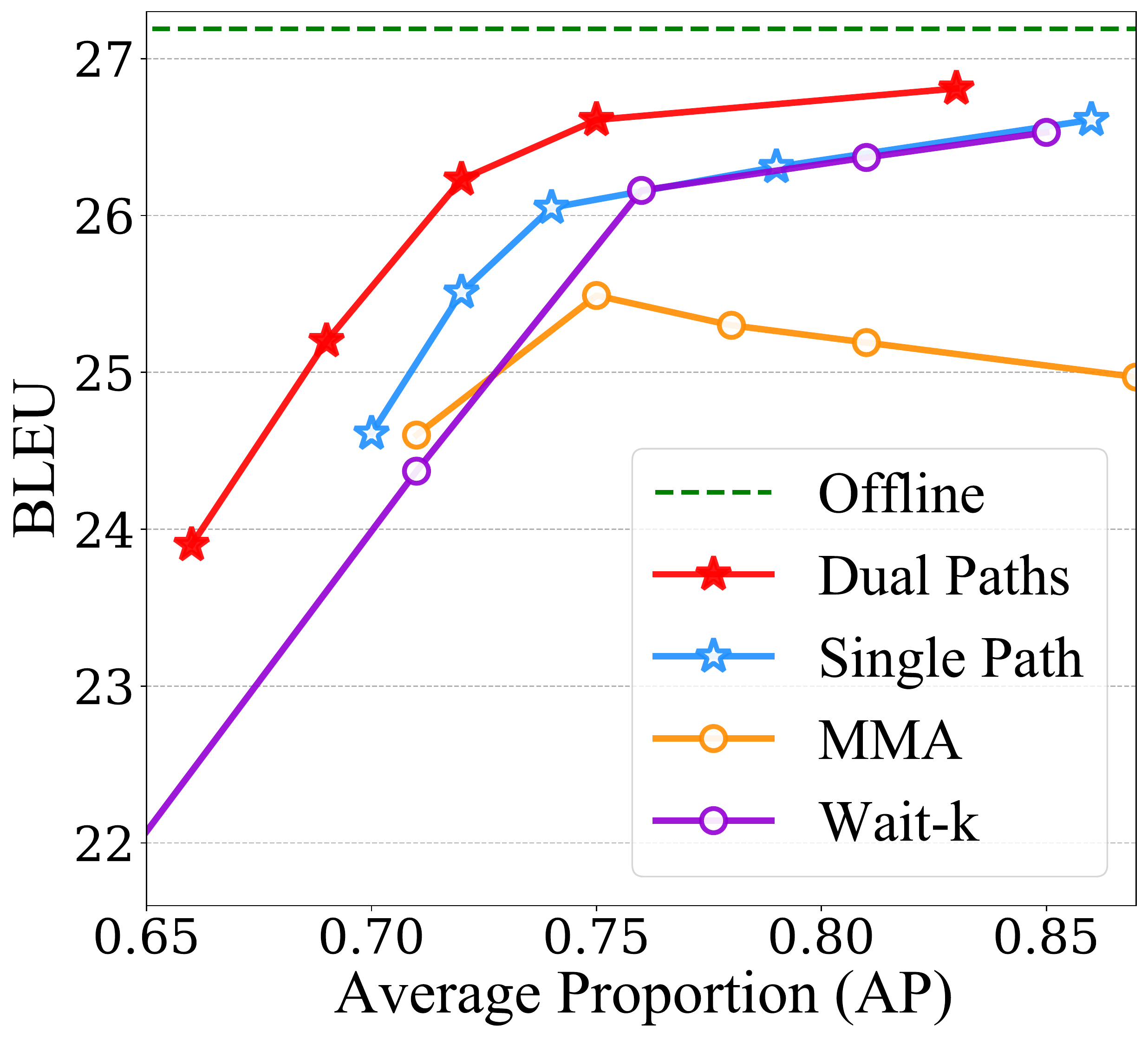}
}
\subfigure[WMT15 En$\rightarrow $De, DAL]{
\includegraphics[width=3in]{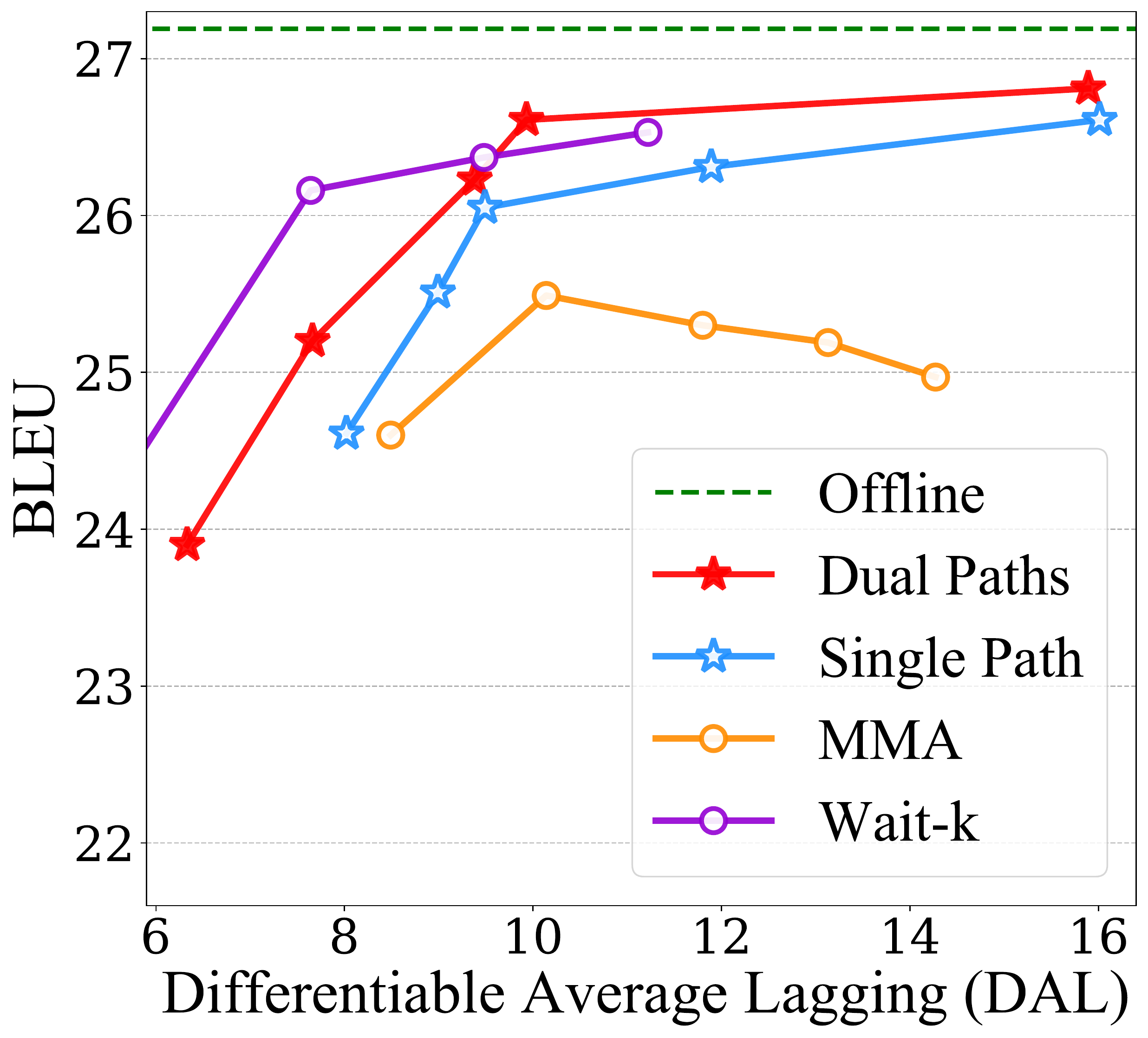}
}

\caption{Results on WMT15 En$\rightarrow $De, measured with AP and DAL.}
\label{endemain}
\end{figure*}

Figure \ref{envimain}, \ref{vienmain}, \ref{deenmain}, \ref{endemain} respectively show the expanded results on IWSLT15 En$\leftrightarrow $Vi and WMT15 De$\rightarrow $En, measured by AP and DAL.

\subsection{Numerical Results}
Table \ref{envi}, \ref{vien}, \ref{deen}, \ref{ende} respectively report the numerical results on IWSLT15 En$\leftrightarrow $Vi and WMT15 De$\rightarrow $En, measured by AP, AL, DAL, and BLEU.

\begin{table}[]
\centering
\begin{tabular}{cC{1cm}C{1cm}C{1cm}C{1.2cm}} \hlinew{1.5pt}
\multicolumn{5}{c}{\textbf{IWSLT15 En$\rightarrow $Vi}}                \\\hlinew{1pt}
\multicolumn{5}{c}{\textit{\textbf{Offline}}}     \\\hline
          & AP      & AL       & DAL     & BLEU   \\
          & 1.00    & $\!\!\!$22.08    & $\!\!\!$22.08   & 28.91  \\\hlinew{1pt}
\multicolumn{5}{c}{\textit{\textbf{Wait-k}}}      \\\hline
$k$         & AP      & AL       & DAL     & BLEU   \\
1         & 0.63    & 3.03     & 3.54    & 25.31  \\
3         & 0.71    & 4.80     & 5.42    & 27.65  \\
5         & 0.78    & 6.46     & 7.06    & 28.54  \\
7         & 0.83    & 8.21     & 8.79    & 28.70  \\
9         & 0.88    & 9.92     & $\!\!\!$10.51   & 28.86  \\\hlinew{1pt}
\multicolumn{5}{c}{\textit{\textbf{MMA}}}         \\\hline
$\lambda$    & AP      & AL       & DAL     & BLEU   \\
0.4       & 0.58    & 2.68     & 3.46    & 27.73  \\
0.3       & 0.59    & 2.98     & 3.81    & 27.90   \\
0.2       & 0.63    & 3.57     & 4.44    & 28.47  \\
0.1       & 0.67    & 4.63     & 5.65    & 28.42  \\
0.04      & 0.70     & 5.44     & 6.57    & 28.33  \\
0.02      & 0.76    & 7.09     & 8.29    & 28.28  \\\hlinew{1pt}
\multicolumn{5}{c}{\textit{\textbf{Single Path}}} \\\hline
$\lambda$    & AP      & AL       & DAL     & BLEU   \\
0.5       & 0.64    & 3.02     & 4.73    & 27.98  \\
0.4       & 0.67    & 3.54     & 5.50    & 28.23  \\
0.3       & 0.67    & 3.83     & 5.57    & 28.52  \\
0.2       & 0.69    & 4.05     & 6.03    & 28.74  \\
0.1       & 0.73    & 5.08     & 7.27    & 28.75  \\
0.05      & 0.75    & 5.38     & 8.14    & 28.76  \\
0.01      & 0.85    & 8.72     & $\!\!\!$12.13   & 28.56  \\\hlinew{1pt}
\multicolumn{5}{c}{\textit{\textbf{Dual Paths}}}  \\\hline
$\lambda$    & AP      & AL       & DAL     & BLEU   \\
0.4       & 0.64    & 3.07     & 4.82    & 28.84  \\
0.3       & 0.66    & 3.49     & 5.46    & 28.68  \\
0.2       & 0.68    & 3.84     & 5.81    & 28.83  \\
0.1       & 0.72    & 4.78     & 7.11    & 28.76  \\
0.05      & 0.78    & 6.14     & 8.93    & 28.87  \\
0.01      & 0.84    & 8.15     & $\!\!\!$11.40   & 29.03 \\\hlinew{1.5pt}
\end{tabular}
\caption{Numerical results of IWSLT15 En$\rightarrow $Vi.}
\label{envi}
\end{table}

\begin{table}[]
\centering
\begin{tabular}{cC{1cm}C{1cm}C{1cm}C{1.2cm}}\hlinew{1.5pt}
\multicolumn{5}{c}{\textbf{IWSLT15 Vi$\rightarrow $En}}                \\\hlinew{1pt}
\multicolumn{5}{c}{\textit{\textbf{Offline}}}     \\
          & AP      & AL       & DAL     & BLEU   \\
          & 1.00    & $\!\!\!$27.56    & $\!\!\!$27.56   & 26.11  \\\hlinew{1pt}
\multicolumn{5}{c}{\textit{\textbf{Wait-k}}}      \\\hline
$k$       & AP      & AL       & DAL     & BLEU   \\
1         & 0.42    & -2.89    & 1.62    & 7.57   \\
3         & 0.53    & -0.18    & 3.24    & 14.66  \\
5         & 0.61    & 1.49     & 5.08    & 17.44  \\
7         & 0.67    & 3.28     & 7.05    & 19.02  \\
9         & 0.76    & 6.75     & 8.96    & 22.39  \\
11        & 0.80    & 7.91     & $\!\!\!$10.71   & 23.28  \\
13        & 0.84    & $\!\!\!$10.37    & $\!\!\!$12.36   & 24.80  \\\hlinew{1pt}
\multicolumn{5}{c}{\textit{\textbf{MMA}}}         \\\hline
$\lambda$    & AP      & AL       & DAL     & BLEU   \\
0.4       & 0.65    & 4.26     & 6.96    & 22.08  \\
0.3       & 0.67    & 4.56     & 7.59    & 22.98  \\
0.2       & 0.69    & 5.03     & 9.09    & 23.50  \\
0.1       & 0.73    & 5.70     & $\!\!\!$11.60   & 24.15  \\
0.05      & 0.80    & 7.51     & $\!\!\!$15.70   & 24.26  \\
0.01      & 0.95    & $\!\!\!$15.55    & $\!\!\!$23.95   & 25.04  \\\hlinew{1pt}
\multicolumn{5}{c}{\textit{\textbf{Single Path}}} \\\hline
$\lambda$    & AP      & AL       & DAL     & BLEU   \\
0.4       & 0.64    & 3.87     & 5.75    & 24.22  \\
0.3       & 0.65    & 4.07     & 5.97    & 24.59  \\
0.2       & 0.67    & 4.55     & 6.70    & 24.63  \\
0.1       & 0.70    & 5.48     & 7.97    & 24.87  \\
0.05      & 0.73    & 6.33     & 9.33    & 25.10  \\
0.01      & 0.84    & $\!\!\!$10.24    & $\!\!\!$13.72   & 25.75  \\\hlinew{1pt}
\multicolumn{5}{c}{\textit{\textbf{Dual Paths}}}  \\\hline
$\lambda$    & AP      & AL       & DAL     & BLEU   \\
0.4       & 0.63    & 3.60     & 5.42    & 24.67  \\
0.3       & 0.66    & 4.52     & 6.63    & 25.28  \\
0.2       & 0.68    & 4.89     & 6.97    & 25.60  \\
0.1       & 0.70    & 5.54     & 8.02    & 26.23  \\
0.05      & 0.75    & 6.95     & $\!\!\!$10.07   & 25.89  \\
0.01      & 0.85    & $\!\!\!$11.30    & $\!\!\!$15.30   & 26.18 \\\hlinew{1.5pt}
\end{tabular}
\caption{Numerical results of IWSLT15 Vi$\rightarrow $En.}
\label{vien}
\end{table}

\begin{table}[]
\centering
\begin{tabular}{cC{1cm}C{1cm}C{1cm}C{1.2cm}} \hlinew{1.5pt}
\multicolumn{5}{c}{\textbf{WMT15 De$\rightarrow $En}}                \\\hlinew{1pt}
\multicolumn{5}{c}{\textit{\textbf{Offline}}}     \\\hline
          & AP      & AL       & DAL     & BLEU   \\
          & 1.00    & $\!\!\!$27.77    & $\!\!\!$27.77   & 30.52  \\\hlinew{1pt}
\multicolumn{5}{c}{\textit{\textbf{Wait-k}}}      \\\hline
$k$         & AP      & AL       & DAL     & BLEU   \\
1         & 0.52    & 0.02     & 1.84    & 16.95  \\
3         & 0.59    & 1.73     & 3.34    & 23.04  \\
5         & 0.66    & 3.86     & 5.20    & 26.06  \\
7         & 0.73    & 5.86     & 7.12    & 27.42  \\
9         & 0.78    & 7.85     & 9.01    & 28.27  \\
11        & 0.82    & 9.75     & $\!\!\!$10.79   & 29.04  \\\hlinew{1pt}
\multicolumn{5}{c}{\textit{\textbf{MMA}}}         \\\hline
$\lambda$    & AP      & AL       & DAL     & BLEU   \\
0.4       & 0.68    & 4.97     & 7.51    & 27.30  \\
0.3       & 0.72    & 6.00     & 9.30    & 28.09  \\
0.25      & 0.78    & 8.03     & $\!\!\!$12.28   & 28.82  \\
0.2       & 0.83    & 9.98     & $\!\!\!$14.86   & 28.79  \\
0.1       & 0.87    & $\!\!\!$13.25    & $\!\!\!$16.48   & 28.39  \\\hlinew{1pt}
\multicolumn{5}{c}{\textit{\textbf{Single Path}}} \\\hline
$\lambda$    & AP      & AL       & DAL     & BLEU   \\
0.4       & 0.60    & 2.73     & 4.73    & 24.93  \\
0.3       & 0.68    & 5.34     & 9.04    & 26.67  \\
0.25      & 0.73    & 6.66     & $\!\!\!$10.08   & 28.19  \\
0.2       & 0.76    & 8.31     & $\!\!\!$12.77   & 28.12  \\
0.1       & 0.86    & $\!\!\!$13.93    & $\!\!\!$17.62   & 29.89  \\\hlinew{1pt}
\multicolumn{5}{c}{\textit{\textbf{Dual Paths}}}  \\\hline
$\lambda$    & AP      & AL       & DAL     & BLEU   \\
0.4       & 0.60    & 2.80     & 4.75    & 26.21  \\
0.3       & 0.62    & 3.19     & 5.40    & 27.04  \\
0.25      & 0.65    & 4.02     & 6.65    & 28.14  \\
0.2       & 0.75    & 7.69     & $\!\!\!$11.51   & 29.23  \\
0.1       & 0.85    & $\!\!\!$13.50    & $\!\!\!$17.59   & 30.10 \\\hlinew{1.5pt}
\end{tabular}
\caption{Numerical results of WMT15 De$\rightarrow $En.}
\label{deen}
\end{table}

\begin{table}[]
\centering
\begin{tabular}{cC{1cm}C{1cm}C{1cm}C{1.2cm}} \hlinew{1.5pt}
\multicolumn{5}{c}{\textbf{WMT15 En$\rightarrow $De}}                \\\hlinew{1.5pt}
\multicolumn{5}{c}{\textit{\textbf{Offline}}}     \\\hline
          & AP      & AL       & DAL     & BLEU   \\
          & 1.00    & $\!\!\!$26.56    & $\!\!\!$26.56   & 27.19  \\\hlinew{1.5pt}
\multicolumn{5}{c}{\textit{\textbf{Wait-k}}}      \\\hline
$k$         & AP      & AL       & DAL     & BLEU   \\
1         & 0.56    & 1.52     & 2.38    & 16.72  \\
3         & 0.64    & 3.46     & 3.97    & 21.69  \\
5         & 0.71    & 5.25     & 5.72    & 24.37  \\
7         & 0.76    & 7.14     & 7.64    & 26.16  \\
9         & 0.81    & 8.96     & 9.48    & 26.37  \\
11        & 0.85    & $\!\!\!$10.76    & $\!\!\!$11.22   & 26.53  \\\hlinew{1.5pt}
\multicolumn{5}{c}{\textit{\textbf{MMA}}}         \\\hline
$\lambda$    & AP      & AL       & DAL     & BLEU   \\
0.4       & 0.71    & 5.54     & 8.49    & 24.60  \\
0.3       & 0.75    & 6.69     & 10.14   & 25.49  \\
0.25      & 0.78    & 7.40     & $\!\!\!$11.80    & 25.30  \\
0.2       & 0.81    & 8.64     & $\!\!\!$13.13   & 25.19  \\
0.1       & 0.87    & $\!\!\!$11.12    & $\!\!\!$14.27   & 24.97  \\\hlinew{1.5pt}
\multicolumn{5}{c}{\textit{\textbf{Single Path}}} \\\hline
$\lambda$    & AP      & AL       & DAL     & BLEU   \\
0.4       & 0.70    & 5.19     & 8.02    & 24.61  \\
0.3       & 0.72    & 5.73     & 8.99    & 25.51  \\
0.25      & 0.74    & 6.39     & 9.49    & 26.05  \\
0.2       & 0.79    & 8.11     & $\!\!\!$11.89   & 26.31  \\
0.1       & 0.86    & $\!\!\!$11.93    & $\!\!\!$16.01   & 26.61  \\\hlinew{1pt}
\multicolumn{5}{c}{\textit{\textbf{Dual Paths}}}  \\\hline
$\lambda$    & AP      & AL       & DAL     & BLEU   \\
0.4       & 0.66    & 4.24     & 6.33    & 23.90  \\
0.3       & 0.69    & 4.80     & 7.66    & 25.20  \\
0.25      & 0.72    & 5.95     & 9.38    & 26.23  \\
0.2       & 0.75    & 6.42     & 9.93    & 26.61  \\
0.1       & 0.83    & $\!\!\!$11.80    & $\!\!\!$15.89   & 26.81  \\\hlinew{1.5pt}
\end{tabular}
\caption{Numerical results of WMT15 En$\rightarrow $De.}
\label{ende}
\end{table}

\end{document}